\documentclass[10pt, two column, twoside]{IEEEtran}

\usepackage{color}
\usepackage{xcolor}
\usepackage{cite}
\RequirePackage{algorithm, algorithmicx, algpseudocode}

\usepackage{amsmath,amssymb,amsthm, amsfonts}
\usepackage{bm} 
\usepackage{mathrsfs}
\usepackage{subfigure}
\usepackage{graphicx,}
\usepackage{epstopdf}
\graphicspath{{figures/}}

\definecolor{colorhkustblue}{RGB}{20,43,140}
\definecolor{colorsinghuapurple}{RGB}{116,52,129}
\definecolor{colorshanghaitechred}{RGB}{164,0,6}
\definecolor{colordeepred}{RGB}{128,0,0}

\newcommand{\reviseZP}[1]{\textcolor{black}{#1}}

\theoremstyle{plain}

\theoremstyle{definition}

\usepackage{array}
\usepackage{booktabs}
\usepackage{multirow}

\begin{document}

\title{Trustworthy Federated Learning via Blockchain }
\author{ Zhanpeng~Yang, \textit{Student Member}, Yuanming~Shi, \textit{Senior Member}, \textit{IEEE}, Yong~Zhou, \textit{ Member}, \textit{IEEE},    Zixin~Wang, \textit{Student Member}, \textit{IEEE}, Kai~Yang, \textit{Member}, \textit{IEEE}
	\thanks{
		Zhanpeng Yang, Yuanming Shi, Yong Zhou,  and Zixin~Wang are with the School of Information Science and
		Technology, ShanghaiTech University, Shanghai 201210, China (e-mail:
		yangzhp@shanghaitech.edu.cn; shiym@shanghaitech.edu.cn; zhouyong@shanghaitech.edu.cn; wangzx2@sha\\nghaitech.edu.cn).
	}
	\thanks{Kai Yang is with JD Technology Group, Beijing 100176, China (e-mail: yangkai188@jd.com).}
}

\maketitle
\begin{abstract}
	The safety-critical scenarios of artificial intelligence (AI), such as autonomous driving, Internet of Things, smart healthcare, etc., have raised critical requirements of trustworthy AI to guarantee the privacy and security with reliable decisions. As a nascent branch for trustworthy AI, federated learning (FL) has been regarded as a promising privacy preserving framework for training a global AI model over collaborative devices. However, security challenges still exist in the FL framework, e.g., Byzantine attacks from malicious devices, and model tampering attacks from malicious server, which will degrade or destroy the accuracy of trained global AI model. In this paper, we shall propose a decentralized blockchain based FL (B-FL) architecture by using a secure global aggregation algorithm to resist malicious devices, and deploying practical Byzantine fault tolerance consensus protocol with high effectiveness and low energy consumption among multiple edge servers to prevent model tampering from the malicious server. However, to implement B-FL system at the network edge, multiple rounds of cross-validation in blockchain consensus protocol will induce long training latency. We thus formulate a network optimization problem that jointly considers bandwidth and power allocation for the minimization of long-term average training latency consisting of progressive learning rounds. We further propose to transform the network optimization problem as a Markov decision process and leverage the deep reinforcement learning based algorithm to provide high system performance with low computational complexity. Simulation results demonstrate that B-FL can resist malicious attacks from edge devices and servers, and the training latency of B-FL  can be significantly reduced by deep reinforcement learning based algorithm compared with baseline algorithms.
\end{abstract}


\section{Introduction}

Artificial Intelligence (AI) has yielded a bloom of worldwide developments and profoundly changed human life, such as autonomous driving \cite{zhang2020mobile}, Internet of Things (IoT) \cite{yang2020federated}, smart healthcare  \cite{hayyolalam2021edge}, etc. However, recent research results have found that AI may cause potential vulnerabilities by, for example, leaking privacy data or making unreliable decisions under adversarial attacks (e.g., misdiagnosed disease in smart healthcare). Consequently, a new AI paradigm is emerging, named \textit{trustworthy AI} \cite{liu2021trustworthy},  which aims at avoiding unfavorable impacts from AI and has garnered increased interest in both academia and industry. In particular, the trustworthy AI intends to achieve stable and sustained high learning accuracy under the considerations of robustness, privacy, accountability, fairness, interpretability, and environmental well-being \cite{liu2021trustworthy}.
As a promising framework of trustworthy AI, federated learning (FL) has recently been proposed to guarantee data privacy in the AI model training process, which collaboratively trains a global AI model by coordinating multiple devices \cite{yang2020federateda,lim2021dynamic}. In the server-client architecture of FL, each device executes local training and uploads its updated local model to a server without transmitting the private raw data. The server then aggregates the collected local models for global AI model update, followed by disseminating the updated global model to devices \cite{yang2020federated}.
Therefore, FL provides a privacy protection solution for privacy-sensitive intelligent applications, e.g., smart healthcare, financial industry, and IoT  \cite{letaief2021edge,shi2020communicationefficient,wang2022federated,yin2022flight,liu2022joint}.

However, security challenges still exist in FL due to the potential malicious devices and malicious server, which will degrade the performance of AI model training or destroy the FL training process \cite{letaief2021edge}. Although FL guarantees the privacy protection of edge devices, the global model can be attacked by malicious edge devices (e.g., model poisoning or adversarial attacks) \cite{so2021byzantineresilient}. Various secure model aggregation algorithms have been proposed to address this challenge, e.g., geometric median, trimmed mean, and multi-Krum \cite{blanchard2017machine}.
In particular, the server-client architecture is vulnerable to server’s malfunction, which includes a potential single point of failure or tamper of global model. The failure or tamper in the server will collapse the entire FL training. To address this issue, blockchain was adopted to establish a decentralized blockchain-based FL (B-FL) network \cite{guo2020adaptive,xiong2019cloud,kang2019secure}, which leverages multiple servers to execute global model aggregation and confirm the correctness of global model using consensus protocols (e.g., Proof of Work (PoW), Proof of Stake (PoS), Raft, and practical Byzantine fault tolerance (PBFT)) \cite{xiong2018when,xiao2020survey}.
The B-FL can thus resist failures or attacks of servers and devices by building trustworthy global model aggregation with secure model aggregation based on blockchain consensus protocol among multiple servers \cite{warnat-herresthal2021swarm,nguyen2021federated, wang2021blockchainbased,zhou2020pirate}.

Moreover, the latency of FL training process, one of key metrics in the edge AI, is also crucial to improve the communication efficiency of FL due to the fading nature of wireless channels \cite{letaief2019roadmap,shi2021mobile,chang20226genabled,yang2021communicationefficient}.
To implement the FL over wireless networks, the edge devices transmit their local models and the edge server disseminates the shared global model over wireless links. This typically consists of computation latency and communication latency, for which theoretical analysis and resource allocation for FL latency have recently been provided \cite{li2021delay,chen2021convergence,ren2021accelerating,yang2020delay}. Specifically, Li \textit{et al.} \cite{li2021delay} characterized the delay distribution for FL over arbitrary fading channels via the saddle point approximation method and large deviation theory. Chen \textit{et al.} \cite{chen2021convergence} proposed a probabilistic device scheduling policy to minimize the overall training time for wireless FL. Ren \textit{et al.} \cite{ren2021accelerating} formulated a training acceleration optimization problem developed the close-form expressions for joint batch size selection and communication resource allocation. Yang \textit{et al.} \cite{yang2020delay} considered a delay minimization problem to obtain the optimal solution by a bisection search algorithm.
However, the latency in wireless B-FL becomes much more complicated due to the additional multiple rounds of cross-validation among edge servers in blockchain consensus protocol. The computation latency includes the local training, global model aggregation and block validation, which depends on the computation capability of edge devices and servers, and the size of dataset and models. The communication latency includes the uplink and downlink transmissions for model updates and blockchain consensus protocol, which mainly depend on wireless communication techniques, bandwidth, and power budgets. It is thus critical to characterize the latency of the wireless B-FL system and optimize the network resources to reduce the overall learning latency.

In this paper, we focus on designing a B-FL architecture to support trustworthy and low latency AI service under the consensus protocol of blockchain. Specifically, we utilize the PBFT consensus protocol  \cite{castro1999practical} to achieve high effectiveness and low energy consumption compared with the existing consensus protocols such as proof-of-work (PoW), and Raft with some loss of security \cite{xiao2020survey}.  The training latency in B-FL system is characterized by considering the computation latency and communication latency, which consists of local training, global model aggregation, consensus protocol, and model dissemination. We then propose to minimize the long-term average learning latency by allocating bandwidth and power resources. This long-term resource allocation problem cannot be equivalently transformed to multiple one-shot problems with respect to the channel of each round, due to the correlation in the long-term average power constraint \cite{shen2020lorm}. Besides, as the optimization variables are coupled in the objective function and constraints, the conventional optimization-based algorithms have prohibitive computational complexity and may be inapplicable in the time-sensitive B-FL system \cite{wen2020joint,xiong2021uavassisted,xiong2019cloud}.
To address the challenges, we propose to transform the long-term resource allocation problem to a Markov decision process (MDP), for which a deep reinforcement learning (DRL) based algorithm is developed to achieve efficient and adaptive resource allocation with low computational complexity from a long-term perspective. The developed DRL-based algorithm will minimize the cumulative latency from a long-term perspective, which balances the latency in current round and that in the future rounds by dynamic resource allocation. Moreover, the DRL-based algorithm establishes a direct mapping from the current network information to resource allocation by deep neural network (DNN), which can significantly reduce the computational complexity.

\subsection{Contributions}
In this paper, we propose a PBFT-based wireless B-FL architecture to realize trustworthy AI training process. To improve the communication efficiency of wireless B-FL, we further propose to minimize long-term average training latency by allocating the bandwidth and power resources. The long-term resource allocation problem is modeled as an MDP, followed by developing a DRL-based algorithm to achieve high system performance and reduce computational complexity. The major contributions of this paper are summarized as follows
\begin{enumerate}
	\item We proposed a PBFT-based wireless B-FL architecture by combining the permissioned blockchain and wireless FL system at the network edge, which builds a trustworthy AI model training environment to resist the failures and attacks from malicious edge servers and malicious edge devices. The wireless B-FL utilizes PBFT consensus protocol of blockchain to validate the correctness of global model update, which achieves high effectiveness and low energy consumption. The detailed procedures of PBFT based wireless B-FL system are presented, followed by characterizing the training latency by considering the communication and computation processes.

	\item By jointly considering the consensus and training processes in the long-term process of B-FL, we formulate a long-term average latency minimization problem by optimizing bandwidth allocation and power allocation. We adopt long-term average power constraint for dynamic power allocation to assign more power in worse channel conditions and less power in better channel conditions. The long-term resource allocation problem turns out to be highly intractable due to the correlation in the long-term average power constraint, and thus cannot be equivalently transformed to multiple one-shot problems.

	\item To solve the long-term resource allocation problem, we propose to transform the resource allocation problem into an MDP, followed by developing a DRL based algorithm to design efficient and adaptive resource allocation scheme with low computational complexity from a long-term perspective. We then utilize a twin delayed deep deterministic policy gradient (TD3) algorithm in DRL to tackle the continuous optimization variables.

\end{enumerate}

The simulation results demonstrate the effectiveness of wireless B-FL to resist malicious attacks from edge devices and servers, which are simulated in MINST dataset for handwriting recognition and heart activity dataset for affect recognition. We further simulate the proposed DRL-based resource allocation scheme with various system parameters, and demonstrate the effectiveness for achieving significant reduction on training latency compared with other baseline algorithms.

\subsection{Related Works}
\begin{table*}\footnotesize
	\newcommand{\tabincell}[2]{\begin{tabular}{@{}#1@{}}#2.8\end{tabular}}
	\renewcommand\arraystretch{1.2}
	\caption{
		\vspace*{-0.1em}List of Notations.}\vspace*{-0.6em}
	\centering  \label{tab_notations}
	\begin{tabular}{|c||c|c||c|}
		\hline
		\textbf{Notation} & \textbf{Description}                       & \textbf{Notation}             & \textbf{Description}                                                         \\
		\hline
		$M$               & number of edge servers                     & $b^{\max}$                    & maximum system bandwidth                                                     \\
		\hline
		$K$               & number of devices                          & $\bar{p}$                     & long-term average power constraint                                           \\
		\hline
		$\mathcal{B}$     & set of edge servers                        & $h_{D_k,B_m}^t $              & channel gain between edge device $D_k$ and edge server $B_m$ at $t$-th round \\
		\hline
		$\mathcal{D}$     & set of edge devices                        & $ \bm{b}^t  $                 & bandwidth allocation at $t$-th round                                         \\
		\hline
		$s_{D_k}$         & batch size of devices $D_k$                & $\bm{p}_t$                    & transmit power of edge servers and devices at $t$-th round                   \\
		\hline
		$\mathcal{S}$     & set of local datasets                      & $R_{D_k,B_m}  $               & achievable transmission rate from edge device $D_k$ to edge server $B_m$     \\
		\hline
		${B_p}$           & index of primary edge server               & $ f_{B_m} $                   & CPU frequency of edge server $B_m$                                           \\
		\hline
		$\bm{w}_k$        & local model of device $U_k$                & $ \sigma $                    & unit CPU cycle for executing secure model aggregation                        \\
		\hline
		$\bm{w}_g$        & global FL model                            & $ \delta $                    & CPU cycle for training one sample                                            \\
		\hline
		$f $              & number of Byzantine edge servers           & $ \rho $                      & CPU cycle for generating or verifying one signature                          \\
		\hline
		$D(B) $           & digest of block (e.g. hash value)          & $ \varpi $                    & size of transactions                                                         \\
		\hline
		$H_B$             & block height in blockchain                 & $ S_B $                       & size of blocks                                                               \\
		\hline
		$N_0$             & power spectral density of AWGN             & $ S_M $                       & size of consensus messages                                                   \\
		\hline
		$\gamma$          & discount factor                            & $ \phi,\theta_1,\theta_2 $    & parameters of online actor and critic networks                               \\
		\hline
		$\eta_a,\eta_c$   & learning rate of actor and critic networks & $ \phi',\theta_1',\theta_2' $ & parameters of target actor and critic networks                               \\
		\hline
		$\kappa$          & update proportion of target networks       & $ \mathcal{R} $               & reply memory buffer                                                          \\
		\hline
		$\vartheta$       & update frequency of target networks        & $ E $                         & steps of exploration                                                         \\
		\hline
	\end{tabular}
	\vspace{-0.3cm}
\end{table*}

The blockchain-based FL system has recently received significant interests in designing trustworthy AI by leveraging the consensus protocol of blockchain and a recent survey paper has presented the fundamental concepts and opportunities in the integration of FL and blockchain, which is called FLchain \cite{nguyen2021federated}.
In particular,
Qu \textit{et al.} \cite{qu2020decentralized}, and  Ma \textit{et al.} \cite {ma2020when}  proposed a B-FL architecture based on PoW consensus protocol, which uses the miners in the blockchain as the centralized aggregator to replace the single dedicated server in the traditional FL system.
Besides,
Li \textit{et al.} \cite{li2020blockchainbased} proposed a blockchain-based federated learning framework with an innovative committee consensus mechanism to avoid malicious central servers and realize effective decentralized storage.
The Biscotti in \cite{shayan2020biscotti} designed a proof-of-federation (PoF) consensus protocol, which coordinates FL training between devices to generate blockchain by providing beneficial model updates or by facilitating the consensus process.
Lu \textit{et al.} \cite{lu2021lowlatency} proposed a B-FL with delegated proof of stakes (DPoS), which is developed in digital twin wireless networks. Based on the above architectures of B-FL with different consensus protocols, there are several works optimizing the network efficiency, e.g., latency and energy consumption.
Kim \textit{et al.} \cite{kim2020blockchained}, Li \textit{et al.} \cite{li2022blockchain}, and Pokhrel \textit{et al.}, \cite{pokhrel2020federated} analyzed an end-to-end latency model of B-FL based on PoW,  characterized the close-form of optimal block generation rate in different scenarios, and formulated an edge association by jointly considering digital twin association, training data batch size, and bandwidth allocation, respectively. 
The paper \cite{hieu2020resource} considered blockchain-enabled FL with PoW consensus protocol and achieved a certain model accuracy while minimizing the energy consumption and training latency with a reasonable payment by DRL approach.  Nguyen \textit{et al.} \cite{nguyen2022latency} employed a decentralized FL model aggregation algorithm in a peer-to-peer-based blockchain network and formulated a system latency minimization problem by optimizing communication and computation resources. Furthermore, to improve privacy protection
Zhao \textit{et al.} \cite{zhao2021privacypreserving } jointly enforced differential privacy on the extracted features and proposed a new normalization technique to protect customers’ privacy and improve the test accuracy in the B-FL system with Algorand consensus protocol. Moreover, several works are already deployed in the real blockchain platforms. For example, Zhang \textit{et al.} \cite{zhang2021blockchainbased}   tackled the challenge of data heterogeneity in failure detection of industrial IoT and implement the B-FL system in Ethereum to evaluate the feasibility, accuracy, and performance. Kang  \textit{et al.} \cite{kang2019incentive} presented efficient reputation management for mobile devices based on consortium blockchain and establish the reputation blockchain system on the Corda.

However, all the above works simply adopted the traditional PoW consensus protocol \cite{li2022blockchain,ma2020when,kim2020blockchained} or proposed some immature consensus protocols (e.g.,  committee consensus mechanism \cite{li2020blockchainbased}, PoF \cite{shayan2020biscotti}).  In this paper, we utilize the PBFT consensus protocol to build a permissioned blockchain, which achieve high effectiveness and low energy consumption compared with famous and widely-used PoW \cite{xiao2020survey}.

\subsection{Organization}
The rest of the paper is organized as follows. Section \ref{sec_model} introduces the wireless B-FL system architecture and the communication model. Section \ref{sec_latency} analyzes the training latency  of wireless B-FL and formulates the long-term average latency minimization problem. The TD3-based DRL algorithm is developed
in Section \ref{sec_TD3}. Section \ref{sec_results} presents simulation results of the proposed algorithm by comparing with other baseline algorithms. The notations in this paper are listed in Table \ref{tab_notations}.

\section{System Model}\label{sec_model}
In this section, we propose a blockchain-based FL architecture and describe the proposed B-FL system over wireless networks.
The procedures of our proposed wireless B-FL are then provided.

\subsection{Architecture of Wireless Blockchain-Based  FL}

\begin{figure}[t]
	\centering
	\includegraphics[width=0.9\linewidth]{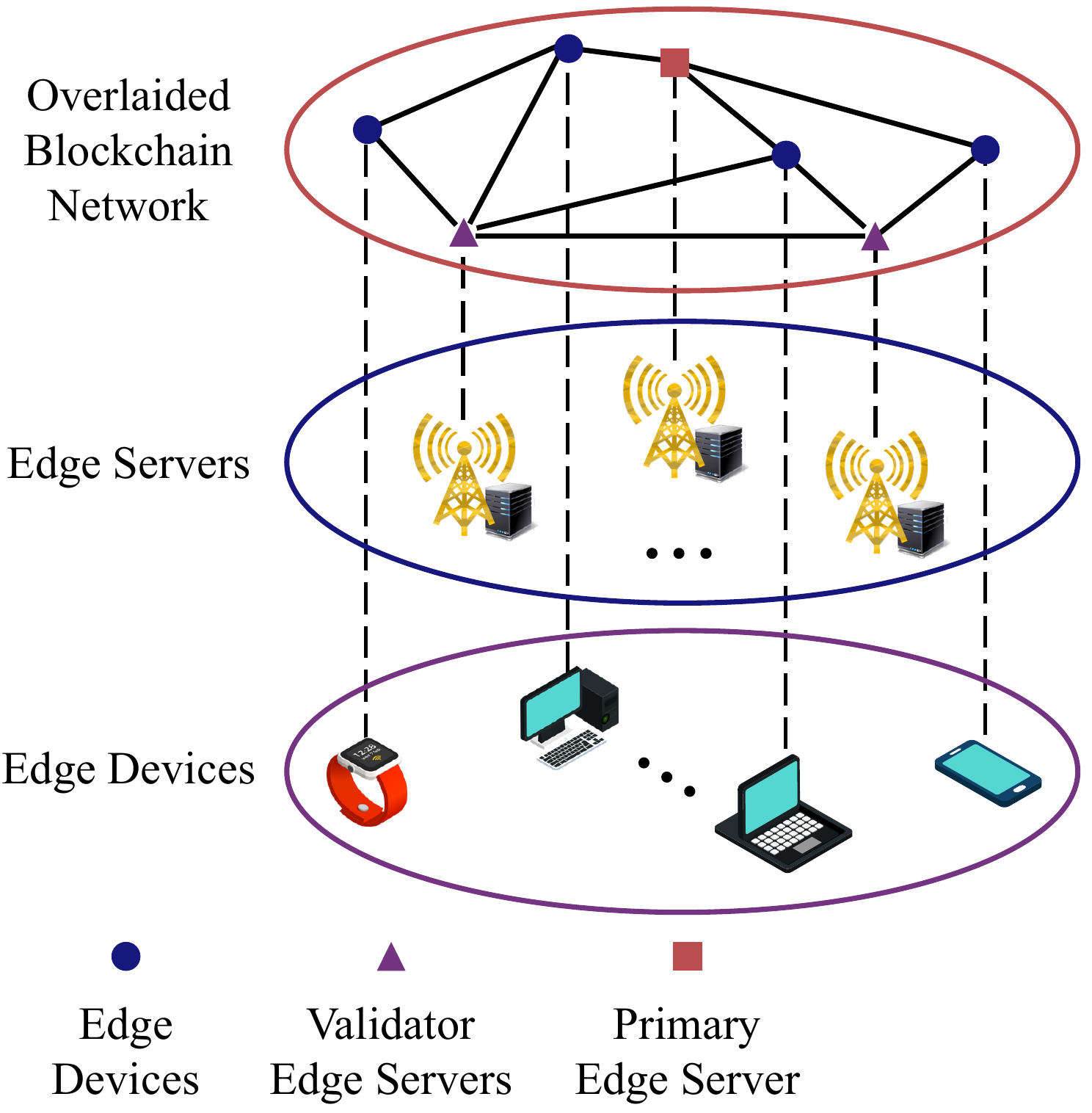} 
	\caption{Wireless blockchain-based federated learning system.}\label{fig_network}
\end{figure}

In this paper, we instead consider a wireless B-FL system consisting of $M$ edge servers and a set of $K$ edge devices, as shown in Fig. \ref{fig_network}. Each edge server is equipped with a mobile edge computing (MEC) sever to execute  computing tasks, and the edge servers are indexed by $\mathcal{B}=\{B_1,\cdots,B_m,\cdots,B_M\}$. Each edge server has enough computation and storage resources to execute global model aggregation and consensus protocol \cite{guo2020adaptive}.  Meanwhile, $K$ edge devices are indexed by $\mathcal{D}=\{D_1,\cdots,D_k,\cdots,D_K\}$. The edge devices with local dataset $\mathcal{S}=\{\mathcal{S}_1,\cdots,\mathcal{S}_k,\cdots,\mathcal{S}_K\}$  have enough computing resources and storage resources to execute the local training process of B-FL. The edge servers communicate with each other over wireless channels \cite{guo2020adaptive,jiang2020intelligent}.
In summary, the B-FL  architecture consists of three layers, i.e., the edge device layer to train local model, the edge server layer to execute secure mode aggregation by blockchain technique,  and overlaided blockchain network layer, as depicted in Fig. \ref{fig_network}.

The conventional  server-client architecture for wireless FL consists of single edge server and multiple edge devices, which collaboratively  train a shared global model without sharing private data of edge devices \cite{kairouz2021advances,lim2022decentralized}.
However, the server-client architecture is vulnerable to edge server’s malfunction, which includes a potential single point of failure and malicious global model poison \cite{li2020federated}. A failure in the edge server may collapse the entire FL network.
In this paper, the proposed wireless B-FL combines the wireless FL with  permissioned blockchain to execute global model aggregation based on blockchain consensus protocol among multiple edge servers, which can resist failures or attacks of edge servers for trustworthy global model aggregation \cite{warnat-herresthal2021swarm}.
The permissioned blockchain provides an attractive solution for malicious server attacks in conventional single server wireless FL due to its features such as decentralization, immutability, and traceability \cite{zhou2020pirate,ma2020when}.
In particular, the wireless B-FL system consists of multiple edge servers and multiple edge devices \cite{yang2019integrated}, which generates a blockchain in a form of blocks linked by cryptography under the control of a consensus protocol to confirm that the data in the block is correct and immutable.
The block is a data structure consisting of all information of local models and global model.
Therefore, the decentralized blockchain technology is adopted to provide a trustworthy model aggregation platform powered by distributed consensus protocol, which mitigates the single point of failure and malicious poison of edge server and ensures transparency and immutable.

Specifically, the B-FL system comprises one primary edge server and several validator edge servers to produce a new block and reach a consensus.  In each training round, each edge device trains the local model based on its own dataset followed by uploading the local model to the primary edge server.  The primary edge server validates the identities of local models and aggregates them to the global model by executing the smart contract. The smart contract consists of lines of code, which is immutable and trackable to execute a function, e.g, secure global model aggregation.  Then, the local models and global model are packed into a new block, which is broadcasted among validator edge servers to validate the correctness of global model.   \reviseZP{ For example, the validator edge servers recalculate the global model and compare it with the global model from the primary edge server. The validator edge servers will reach a consensus on the correctness of global using consensus protocols (e.g., PoW, PBFT). B-FL will enter the next training round if the global model is correct.  However,  if the primary edge  server computes incorrectly or tampers with the global model, the validator edge servers will ignore the global model from the malicious primary edge server and choose another primary edge server to restart the training round.}  After the validator edge servers reaching a consensus, the new block is stored in the blockchain to ensures transparency and traceability. The validated global model is then disseminated to edge devices. Therefore, the consensus protocol is the critical factor to ensure the uniformity and security of the decentralized blockchain system.

In this paper, we adopt the PBFT consensus protocol in the proposed B-FL system to realize secure global model aggregation among edge servers.
Compared with famous and widely-used PoW consensus protocol \cite{qu2020decentralized,ma2020when,kim2020blockchained,li2022blockchain,pokhrel2020federated,hieu2020resource}, PBFT has weaker security performance. Generally, PoW is a permissionless consensus protocol, which means arbitrary edge sever can participate in the consensus and can tolerate 50\% computing power attack.  PBFT is a permissioned consensus protocol, which means only the authorized edge sever can participate in the consensus and can just resist 33\% malicious edge servers.
However, PBFT achieves more effective consensus and lower electrical energy consumption than PoW. In particular, the consensus rate of PBFT is hundreds of times higher than that of PoW and the PoW-based Bitcoin costs as much electrical energy as the whole of Switzerland \cite{xiao2020survey}.  To achieve high effectiveness and low energy consumption, we choose the PBFT consensus protocol in our proposed B-FL system.

\subsection{Procedures of Wireless B-FL}
\label{sec_procedures}

\begin{figure*}[t]
	\centering
	\includegraphics[width=0.9\linewidth]{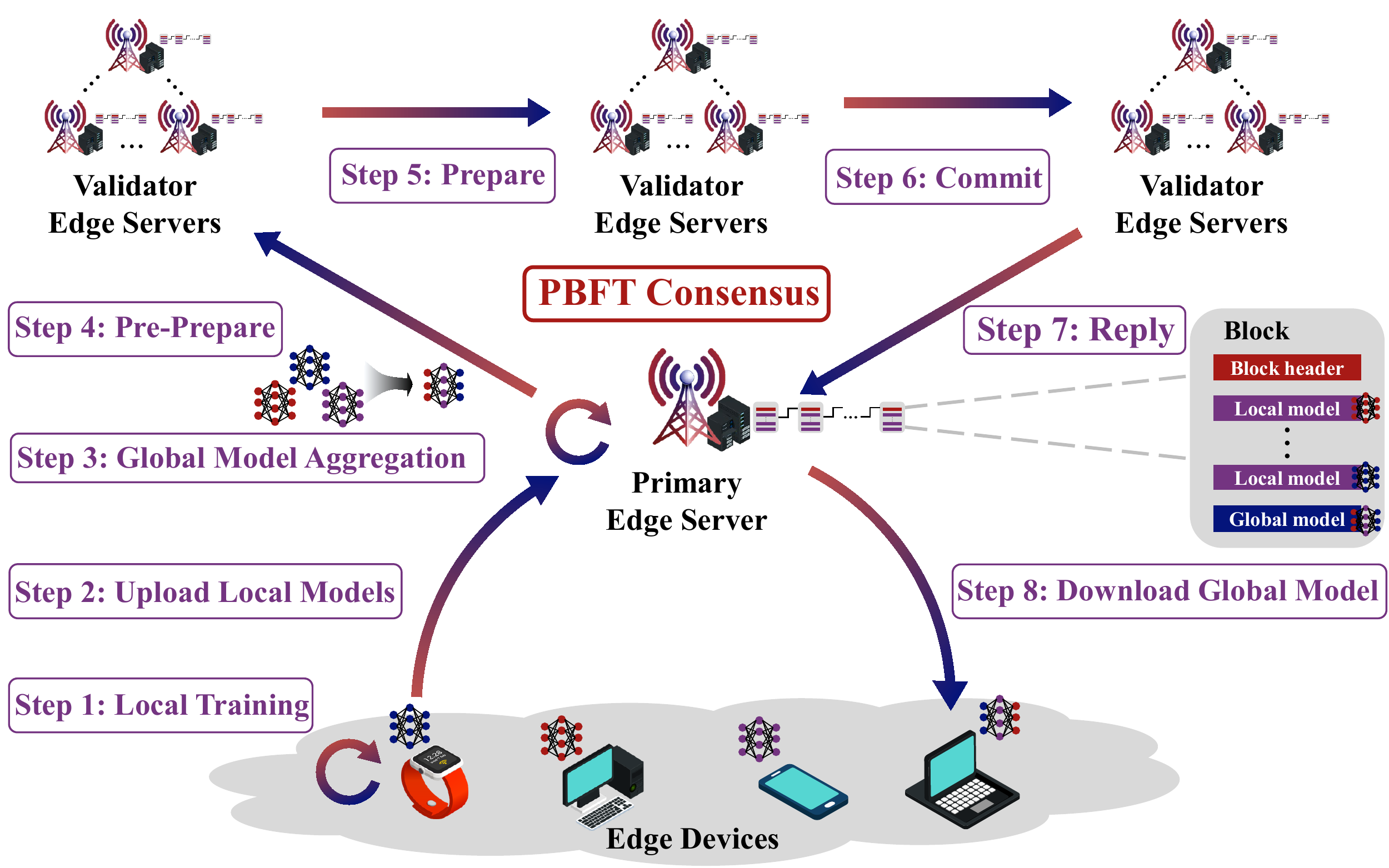} 
	\caption{The procedures of wireless B-FL.}\label{fig_procedure}
\end{figure*}

To train the ML models in the wireless B-FL system based on PBFT consensus protocol, eight steps are required in each round to achieve trustworthy global model aggregation to resist failures and attacks of malicious edge servers. The eight steps are also shown in Fig \ref{fig_procedure}.

\begin{algorithm}[t]
	\caption{Pseudocode of Wireless B-FL}
	\label{alg_FL+blockchain}
	\begin{algorithmic}[1]
		\Require {Edge servers set $\mathcal{B}$, edge devices set $\mathcal{D}$, the dataset of edge devices $\mathcal{S}$.}
		\Ensure {The global model $\bm{w}_g$ and a blockchain that contains training information.}

		\State {Initialize the blockchain $\bm{B}$ and global model $\bm{w}_g^0$.}
		\For{each round $t=1,2,\dots$}
		\State {Rotate primary edge server among  edge servers.}
		\State {Allocate bandwidth and power resources.}
		\For{each device $D_k\in \mathcal{D}$}
		\State {$D_k$ executes local training on its local data $\mathcal{S}_k$ to \Statex \qquad \quad  generated local model $\bm{w}^{t}_k$.}
		\State {$D_k$ uploads its local model to primary edge server.}
		\EndFor
		\State {Primary edge server validates the local models \Statex \qquad $\{\bm{w}^{t}_k\}_{U_k\in \mathcal{U}}$   and aggregates them to a global model \Statex \qquad $\bm{w}_g^{t}={\sf{multi\_KRUM}}(\{ \bm{w}^t_k\}_{U_k\in \mathcal{U}})$.}
		\State {Primary edge server packs local models and global \Statex \qquad  model to a new block and broadcasts the block to the \Statex \qquad validator edge servers.}
		\State {Primary edge server and validator edge servers reach \Statex \qquad consensus by   PBFT protocol consensus (including \Statex \qquad pre-prepare,   prepare, commit, reply steps) to  \Statex \qquad guarantee the  correctness of global model  and add \Statex \qquad verified block   to the blockchain.}
		\State {Primary edge server broadcasts global model $\bm{w}_g^{t}$ to \Statex \qquad edge  devices.}
		\EndFor

		\State {}

		\Function{${\sf{multi\_KRUM}}$}{${\bm{w}^t_k}_{D_k\in  \mathcal{D}}$}
		\State {Compute the Euclidean distances \Statex \qquad $s(D_k)=\sum_{D_j\neq D_k} \| \bm{w}_k-\bm{w}_j\|^2, \forall D_k\in  \mathcal{D} $, where \Statex \qquad the  sum runs over the $K-f-2$ closest vectors and \Statex \qquad $f$  is  a specified parameter with respect to Byzantine \Statex \qquad tolerant.}
		\State {Select the $K-f$ lowest distance and corresponding \Statex \qquad local models as $\hat{\mathcal{D}} \subset  \mathcal{D} $.}
		\State {Compute  $\bm{w}_g= \frac{1}{|\mathcal{S} |} \sum_{D_k \in \hat{\mathcal{D}}} \bm{w}_k$ by averaging.}
		\State {\textbf{return} $\bm{w}_g $.}
		\EndFunction

	\end{algorithmic}
\end{algorithm}

\subsubsection{Local Training}
Each edge device uses the local dataset to train the local model through stochastic gradient descent. For each edge device $D_k\in \mathcal{D}$, the local loss function on the  batch of samples $\mathcal{S}_k$ randomly from local dataset can be denoted by
\begin{equation}
	F_k(\bm{w};{\mathcal{S}}_k)=\frac{1}{s_{D_k}} \sum_{(\bm{x}_i,y_i)\in {\mathcal{S}}_k} f(\bm{w};\bm{x}_i,y_i),
\end{equation}
where $s_{D_k}$ denotes the batch size of edge device $D_k$, $\bm{x}_i$ and $y_i$ are features and label respectively, and $\bm{w}$ denotes the parameter of the ML model. In particular, $f(\bm{w};(\bm{x}_i,y_i))$ is the sample-wise loss function, which depends on the ML model. Then, the local model parameter can be updated by the stochastic gradient descent algorithm, i.e.,
\begin{equation}
	\bm{w}_k^t=\bm{w}_k^{t-1}-\eta \nabla F_k(\bm{w}_k^{t-1};\mathcal{S}_k),
\end{equation}
where $\bm{w}_k^t$ denotes the trained model parameters  of edge device $D_k$ in round $t$, $\nabla F_k({\bm{w}_k^{t-1}};\mathcal{S}_k)$ is the gradient of $F_k(\cdot)$ at point $\bm{w}_k^{t-1}$, and $\eta$ is the learning rate. In this step, the computation latency is the local model parameter calculation at the edge devices.

\subsubsection{Upload Local Models}
With allocated wireless resources (e.g., bandwidth, transmit power), the edge devices upload the local models to the primary edge server as transactions $\left<\bm{w}_k,D_k\right>,\forall D_k \in \mathcal{D} $, where $\left<\bm{w}_k,D_k\right>$ denotes a data packet consisting of local model parameter $\bm{w}_k$ and the data packet is signed by edge device $D_k$.
The digital signature is a cryptology technique, which is used to validate the authenticity and integrity of data packets and guarantee that the information in data packet is not be tampered with others. The transaction is a concept in blockchain system involving cryptocurrency, contracts, records or other information to record all behavior in blockchain. In this paper, we assume that the data packets are encoded into digital signals (e.g. polar code) and transmitted by rate adaptation scheme to ensure error-free transmission.      It is worth noting that the primary edge server is appointed  before each training round (e.g., the primary edge server rotates among all the edge servers) and all edge devices and edge servers know the primary edge server before each round.  In this step, the communication latency is the transaction uploading for all edge devices.

\subsubsection{Global Model Aggregation}

After collecting the data packets $\left<\bm{w}_k,D_k \right>,\forall D_k \in \mathcal{D}$, the primary edge server validates the transactions to confirm the local model owner valid and executes smart contract to aggregate them into a global model. The primary edge server aims at minimizing the global loss function through global aggregation, which is defined as
\begin{equation}\label{equ_global_loss}
	F(\bm{w})=\sum_{D_k\in \mathcal{D}} F_k(\bm{w};\mathcal{S}_k).
\end{equation}
Although FL  guarantees the privacy protection of edge devices to a certain extent, the global model can be  attacked by  malicious edge devices (e.g., data or model poisoning) \cite{so2021byzantineresilient}. \reviseZP{The proposed B-FL is compatible with all secure global aggregation algorithms (e.g., geometric median, trimmed mean, and Multi-Krum). We use multi-KRUM algorithm to realize Byzantine-resilient secure global model aggregation in this paper, which outperforms other algorithm in convergence speed} \cite{blanchard2017machine,shayan2020biscotti,huang2021byzantineresilient}, i.e.,
\begin{equation}
	\bm{w}^t_g= {\sf{multi\_KRUM}}(\{\bm{w}^t_k,\forall D_k\in \mathcal{D}\}),
\end{equation}
where $\bm{w}^t_g $ is the global model in round $t$, and $\bm{w}_k^t$ is the local model from edge device $D_k$. The function ${\sf{multi\_KRUM}}(\cdot)$ is the multi-KRUM algorithm, which is defined in Algorithm \ref{alg_FL+blockchain}.
After that, the local models and global model are packed into a new block, which is a data packer denoted by $B=\left<\{\left<\bm{w}_1,D_1\right>\}_{D_k\in \mathcal{D}},\left<\bm{w}_g,B_p\right>\right>$, where $B_p$ is the notation of primary edge server. In this step, the computation latency is the transaction validation and model aggregation in the primary edge server.

\subsubsection{Pre-Prepare}
After producing the new block, the primary edge server broadcasts the new block with a pre-prepare message, i.e., $\left<{\text{PRE-PREPARE}},H_B,D(B),B,B_p\right>$.  $H_B$ is the block height (i.e., the number of valid blocks in the blockchain), $D(B)$ is a digest of $B$ (e.g., hash value) and the pre-prepare message is signed by primary edge server $B_p$. The validator edge servers, which are all edge servers except for the primary edge server, receive the new block and verify the digital signature of transactions and new block to confirm the  valid digital signature of new block. Furthermore, the global model is recalculated to confirm that the primary edge server computes correctly. After verification, each validator edge server stores the new block to local. In this step, the computation latency is the digital signature validation of new block  and recalculation of global model. The communication latency is the block transmission from the primary edge server.

\subsubsection{Prepare}
After verification, the validator edge servers transmit the prepare messages $\left<{\text{PREPARE}},H_B,D(B),B_m\right>$, $\forall B_m\in \mathcal{B},  B_m\neq B_p$ to others. The primary edge server and all validator edge servers receive prepare messages and confirm that a majority of validator edge servers have verified the new block and agree with the new block. In this step, the computation latency is the digital signature validation of prepare messages  in all edge servers. The communication latency is the message transmission from the validator edge servers.

\subsubsection{Commit}
After the agreement of new block, the validator edge servers will broadcast the commit messages $\left< {\text{COMMIT}},  H_B, D(B), B_m \right>$, $\forall B_m\in \mathcal{B}$ to others. The validator edge servers receive enough commit messages and confirm that most edge servers reach a consensus.  In this step, the computation latency is  the digital signature validation of commit messages in all edge servers. The communication latency is the message transmission from all edge servers.

\subsubsection{Reply}
After reaching the consensus, the new block becomes valid and all validator edge servers return the reply message $\left<{\text{REPLY}},H_B,D(B),B_m\right>,\forall B_m\in \mathcal{B},  B_m\neq B_p$ to the primary edge server, which indicates that they have stored the new block in the blockchain ready for the next consensus process. In this step, the computation latency is  the digital signature validation of reply messages in the primary edge server. The communication latency is the message transmission from the validator edge servers.

\subsubsection{Download Global Model}
The primary edge server sends back the validated global model to the edge devices, which promotes the next round of training. In this step, the communication latency is the global model downloading at the edge devices.

The steps of pre-prepare, prepare, commit, and reply are necessary in decentralized blockchain system to reach a consensus when each edge server does not know who is malicious. The consensus protocol ensures the correctness of global model aggregation by identity authentication and verification of calculation results at validator edge servers. The pseudocode of wireless B-FL is presented in Algorithm \ref{alg_FL+blockchain}.

\subsection{Communication Model}

The orthogonal frequency division multiple access (OFDMA) technique is adopted in wireless B-FL system. 
\reviseZP{All devices and edge servers are assumed to have a single antenna for 
simplification \cite{guo2020adaptive, lu2021lowlatency, nguyen2022latency}, which can be generalized to the multi-antenna case by additionally allocating the power for each antenna of each server.} 
We assume that the channel state information(CSI) is available for resource allocation and static in one time-slot but vary from different time-slots. To model the channel gain in each time-slot, the channel gain between edge device $D_k$ and edge server $B_m$ of time-slot $s$  is defined by $h_{D_k,B_m}[s]=\zeta_{D_k,B_m} |g_{D_k,B_m}[s]|^2 $, where $\zeta_{D_k,B_m}$ and $g_{D_k,B_m}[s]$ are  large-scale path loss attenuation and  small-scale block fading between edge device $D_k$ and edge server $B_m$, respectively. In particular, the large-scale path loss attenuation is defined as $\zeta_{D_k,B_m}=d_{D_k,B_m}^{-\alpha}$, where $d_{D_k,B_m}$  is the distance  between edge device $D_k$ and edge server $B_m$ and $\alpha$ is the path loss exponent. Based on the Jakes' fading model \cite{wang2022decentralized},  small-scale block fading $g_{D_k,B_m}[s]$ is modeled as first-order complex Gauss-Markov process, i.e.,
\begin{equation}
	g_{D_k,B_m}[s]=\varrho g_{D_k,B_m}[s-1]+\sqrt{1-\varrho^2} \epsilon_{D_k,B_m}[s],
\end{equation}
where $\varrho=J_0(2\pi f_d T_0)$ denotes  the correlation coefficient in
two consecutive time slots. $\epsilon_{D_k,B_m}[s]$ follows an independent and identically distributed (i.i.d.) circularly symmetric complex Gaussian distribution with unit variance. Here, $J_0(\cdot)$ is the zeroth-order Bessel function of the first kind, $f_d$ represents the maximum Doppler frequency, and $T_0$ is the duration of one time slot.

In fact, the durations of time-slots are very short (e.g., the frame time-slot of LTE protocol is 10 ms), but the duration of one B-FL training round is often in the time-scale of second because of the high computation and communication complexity in model training and consensus protocol. Therefore, each B-FL training round contains multiple time-slots \cite{ren2021accelerating,li2021delay}. Because the  purpose of this paper is minimization total latency from the long-term perspective with enormous rounds, the channel dynamics of multiple time-slots in a single round is negligible.  Thus, we employ the average channel gain instead of the instantaneous ones as the channel gain in one round to simplify the problem, i.e., $h_{D_k,B_m}^t=\frac{1}{S} \sum_{s=1}^S h_{D_k,B_m}[tS+s] $, which is channel gain between edge device $D_k$ and edge server $B_m$ in training round $t$ and $S$ is the total time-slot in one round \cite{ren2021accelerating}.

Thus, the achievable transmission rate from edge device $D_k$ to edge server $B_m$ is denoted by
\begin{equation}
	R_{D_k,B_m}^t= b_{D_k}^t  \log_2 \left(1+\frac{h_{D_k,B_m}^t p_{D_k}^t}{b_{D_k}^t N_0}\right),
\end{equation}
where $b_{D_k}^t$ is transmission bandwidth assigned to edge device $D_k$ at communication round $t$, and $h_{D_k,B_m}^t$ is the channel gain from edge device $D_k$ to edge server $B_m$. Here,  $p_{D_k}^t$ and $N_0$ represent the transmit power in edge device $D_k$ and the power spectral density of additive white Gaussian noise (AWGN), respectively.
In this paper, the maximum system bandwidth is denoted by $b^{\max}$, i.e., $\sum_{d\in \mathcal{D} \cup \mathcal{B}} b_{d}^t \le b^{\max}$, which means the edge servers and edge devices share the same bandwidth of broadband communication \cite{guo2020adaptive}.
Therefore,  the latency of transmitting data packet, whose size is $\varpi$, from device $D_k$ to edge server $B_m$ can be written as
\begin{equation}
	T_{D_k,B_m}=\frac{\varpi}{R_{D_k,B_m}^t}.
\end{equation}
In this paper, we assume that the data packets are encoded into digital signals (e.g. polar code) and transmitted by rate adaptation scheme to ensure error-free transmission.

\section{System Optimization} \label{sec_latency}

In this section, we will analyze the latency of proposed wireless B-FL system, by considering both computation and communication latency. To achieve low latency and high reliability of wireless B-FL, we formulate the latency optimization problem by jointly optimizing bandwidth allocation and power allocation.

\subsection{Latency Analysis for Wireless B-FL}

\begin{figure}[t]
	\centering
	\includegraphics[width=1\linewidth]{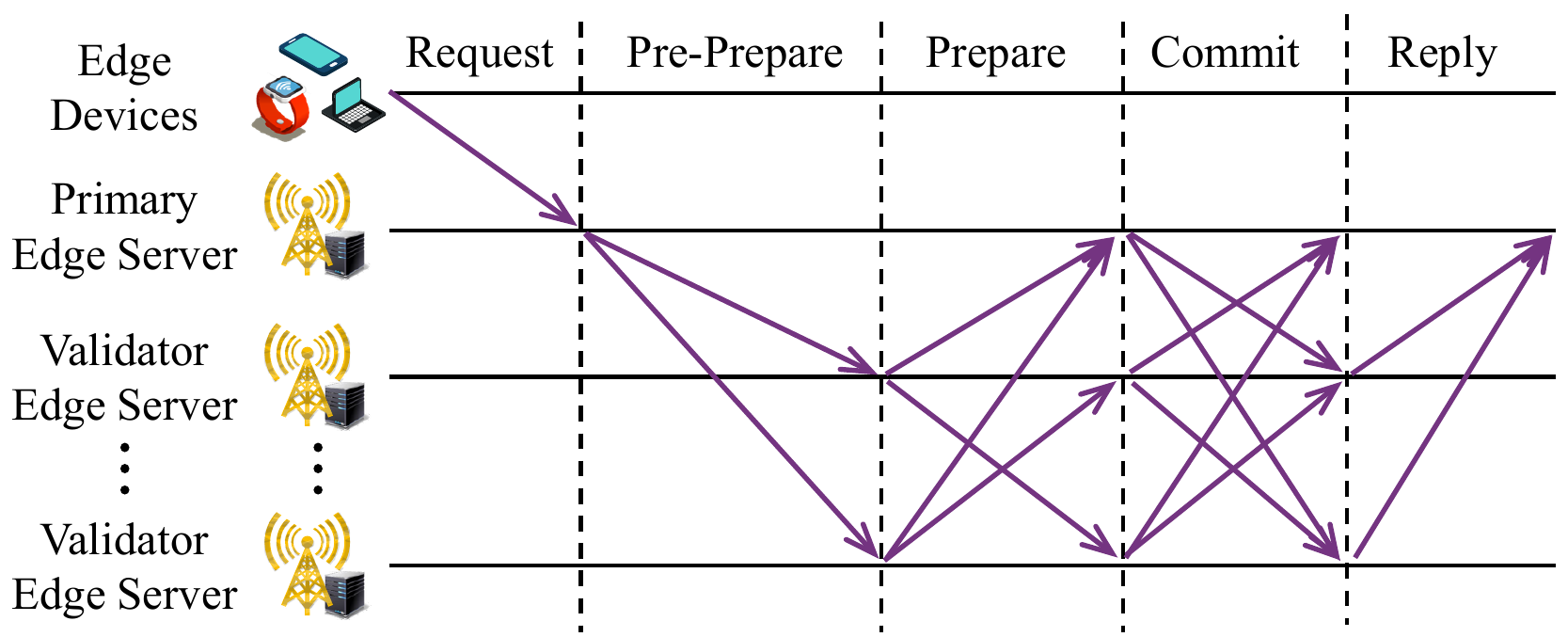} 
	\vspace{-4mm}
	\caption{PBFT consensus protocol. }\label{fig_pbft}
	\vspace{-4mm}
\end{figure}

In the following, we give detailed steps and characterize the  latency of each step in the $t$-th round. We adopt synchronized transmission scheme in this paper to ensure all edge servers and devices in the same step.  Compared with conventional wireless FL system without blockchain, the wireless B-FL system with blockchain needs to validate the digital signature of local model and reaches a consensus of global model aggregation, which can be treated as the extra computation overhead. We assume that  generating or verifying one digital signature requires $\rho$  CPU cycles. The transmission processes of data packets in PBFT consensus protocol are shown in Fig \ref{fig_pbft}, where the request step contains the upload local models step  and global model aggregation step. The latency of eight steps in Section \ref{sec_procedures} is presented as follows.

\subsubsection{Local Training}

The number of CPU cycles for devices to execute backpropagation algorithm with one sample is denoted by $\delta$. The local model weight calculation latency can be expressed as
\begin{equation}
	T^{\text{cmp}}_{\text{train}} = \mathop {\max}\limits_{D_k\in \mathcal{D}} \left\{ \frac{s_{D_k}\delta}{f_{D_k}} \right\},
\end{equation}
where $s_{D_k}$ is the batch size of edge device $D_k$ and $f_{D_k}$ is the CPU frequency of edge device $D_k$.

\subsubsection{Upload Local Models}

After the local training, each updated local model is packed into a transaction, which consists of digital signature and local model. The computation latency of digital signature is
\begin{equation}
	T^{\text{cmp}}_{\text{up}}=\mathop {\max}\limits_{D_k\in \mathcal{D}} \frac{\rho}{f_{D_k}}.
\end{equation}
In this step, we define the average size of one transaction is denoted by $\varpi$, which is related to the AI model size. The transmission latency from edge device to primary edge server can be expressed by
\begin{equation}
	T^{\text{com}}_{\text{up}}=\mathop {\max}\limits_{D_k\in {\mathcal{D}}} \left\{ \frac{\varpi}{R^t_{D_k,B_p}} \right\},
\end{equation}
where $B_p \in \mathcal{B}$ denotes the primary edge server, and $R_{D_k,B_p} $ is the transmission rate from edge device $D_k$ to the primary edge server $B_p$.

\subsubsection{Global Model Aggregation}

The transactions are submitted to the primary edge server. The primary edge server receives these transactions consisting of local models and verifies its digital signature to confirm the identity. Then, the primary edge server will execute smart contract to aggregate local models and generate global model. The smart contract consists of lines of code, which is immutable and trackable to execute a function. Therefore, the secure global model aggregation  via multi-KRUM algorithm can be programmed as a smart contract.  After generating the global model, all the local model transactions and the global model transaction are packed into a new block.

The maximum number of transactions that can be packed in one block is $K+1$, where $K$ is the number of edge devices, i.e., local model number, and one global model transaction.  Thus, the block size can be presented as $S_B=(K+1) \varpi$ without considering the negligible size of block header.
In this step, the primary edge server needs to verify the digital signatures for ${K}$ transactions. Then, the primary edge server  executes the smart contract of secure global model aggregation, whose total CPU cycles can be defined as $\sigma $.  Therefore, the total amount of computation at the primary edge server is $\Delta_{\text{req},B_p}={K}\rho+\sigma$. Then, the computation delay at the primary edge server is

\begin{equation}
	T^{\text{cmp}}_{\text{agg}}=\frac{\Delta_{\text{req},B_p}}{f_{B_p}}.
\end{equation}

\subsubsection{Pre-Prepare}
After generating the new block, the primary edge server will broadcast a pre-prepare message to all the validator edge servers for validation. The pre-prepare message consists of the digital signature of the primary edge server and the information of the new block. The validator edge servers, which are all edge servers except the primary edge server, need to validate that the new block are generated by the given primary edge server and validate the identities of all transactions to ensure the new block is valid. Besides, the validator edge servers are mainly responsible for correctness validation of global model, which can use an  intuitive method by repeatedly calculating the global model through identified local models, and comparing the computation results with the global model of primary edge server.

If the pre-prepare message is validated by $2f$ validator edge servers,
it enters the next step. In particular, $f$ is the number of hypothetical malicious edge servers, which satisfy $3f+1=M$ \cite{castro1999practical}.  PBFT is designed to ensure the correctness  of $2/3$  edge servers, which is more efficient than validating all edge server but less secure. It is a trade-off between efficiency and security.

In this step, the  latency can be presented as
\begin{equation}
	T^{\text{com}}_{\text{prep}}=\mathop {\max}\limits_{B_m\in \mathcal{B}/{B_p}} \left\{ \frac{S_B}{R^t_{B_p,B_m}} \right\}.
\end{equation}
The amount of computation at the validator edge servers is $\Delta_{\text{prep},B_m}=\rho+ (K+1)\rho+\sigma  $, where there is $B_m \neq B_p$. The computation latency in this step is
\begin{equation}
	T^{\text{cmp}}_{\text{prep}}=\mathop {\max}\limits_{B_m\in \mathcal{B}/{B_p}} \left\{ \frac{\Delta_{\text{prep},B_m}}{f_{B_m}} \right\}.
\end{equation}

\subsubsection{Prepare}
After accepting the new block generated by primary edge server, each validator edge server will broadcast a prepare message to all the other edge servers consisting primary edge server and validator edge servers. Each edge server will validate the prepare message to ensure the consistent with the pre-prepare message.

In this step, the communication latency consisting of   broadcasting the prepare message to all other validator edge servers, which is calculated by
\begin{equation}
	T^{\text{com}}_{\text{pre}}=\mathop {\max}\limits_{B_m\in \mathcal{B}/{B_p},B_{m'}\in \mathcal{B},B_m\neq B_{m'}} \left\{ \frac{S_M}{R^t_{B_m,B_{m'}}} \right\},
\end{equation}
where $S_M$ denotes the size of prepare message, which is smaller than $S_B$ because it does not contain the new block.

For the computation, the primary edge server needs to validate $2f$ digital signatures from the other validator edge servers, which can be expressed by $\Delta_{\text{pre},B_p}=2f\rho$. Each of validator edge servers needs to generate one digital signature for the prepare message. Then, $2f$ digital signatures are needed to be validated. Thus, the amount of computation at the validator edge servers $B_m (B_m\neq B_p)$ can be given by $\Delta_{\text{pre},B_m}=\rho+2f\rho$. The computation latency in this step is
\begin{equation}
	T^{\text{cmp}}_{\text{pre}}=\mathop {\max}\limits_{B_m\in \mathcal{B}} \frac{\Delta_{\text{pre},B_m}}{f_{B_m}}.
\end{equation}

\subsubsection{Commit}
After receiving $2f$ valid prepare messages from the other validator edge servers, each edge server will broadcast a commit message to all the other edge servers.
The communication latency can be characterized by
\begin{equation}
	T^{\text{com}}_{\text{cmit}}=\mathop {\max}\limits_{B_m\in \mathcal{B}/{B_p},B_{m'}\in \mathcal{B},B_m\neq,B_{m'}} \left\{ \frac{S_M}{R^t_{B_m,B_{m'}}} \right\}.
\end{equation}

In this step, each edge server needs to generate one digital signature to construct the commit messages. After receiving the commit messages, each edge server needs to verify $2f$ digital signatures. Thus, the amount of computation at each edge server is $\Delta_{\text{cmit},B_m}=\rho+2f\rho$. The computation latency in this step is
\begin{equation}
	T^{\text{cmp}}_{\text{cmit}}=\mathop {\max}\limits_{B_m\in \mathcal{B}} \left\{ \frac{\Delta_{\text{cmit},B_m}}{f_{B_m}} \right\}.
\end{equation}

\subsubsection{Reply}
After collecting $2f$  commit messages, the new block is ensured to a valid block, and it will be appended to the blockchain. A reply message will be transmitted to primary edge server.
In this step, the communication latency is
\begin{equation}
	T^{\text{com}}_{\text{rep}}=\mathop {\max}\limits_{B_m\in \mathcal{B}/{B_p}} \left\{ \frac{S_M}{R^t_{B_m,B_{p}}} \right\}.
\end{equation}

For the computation, each validator edge server needs to generate one digital signatures for the primary edge server, which can be given by $\Delta_{\text{rep},B_m}=\rho$. For the primary edge server, it needs to verify $2f$ signatures, the amount of computation of which is given by $\Delta_{\text{rep},B_p}=2f\rho$. The computation latency in this step is

\begin{equation}
	T^{\text{cmp}}_{\text{rep}}=\mathop {\max}\limits_{B_m\in \mathcal{B}} \left\{ \frac{\Delta_{\text{rep},B_m}}{f_{B_m}} \right\}.
\end{equation}

\subsubsection{Download Global Model}
After local mode validation, global model aggregation, and the new block addition, the global model is transmitted to the edge devices to enter the next round.
In this step, the communication latency is
\begin{equation}
	T^{\text{com}}_{\text{down}}=\mathop {\max}\limits_{D_k\in \mathcal{D}} \left\{ \frac{\varpi}{R^t_{B_p,D_{k}}} \right\}.
\end{equation}

From the above analysis, the total latency in the $t$-th round consists of  communication latency and computation latency, i.e.,
\begin{equation}
	T({\bm{b}^t,\bm{p}^t})=T^{\text{com}}+T^{\text{cmp}}.
\end{equation}
The communication time consumption can be calculated as
\begin{equation}
	T^{\text{com}}=T^{\text{com}}_{\text{up}}+T^{\text{com}}_{\text{prep}}+T^{\text{com}}_{\text{pre}}+T^{\text{com}}_{\text{cmit}} +T^{\text{com}}_{\text{rep}}+T^{\text{com}}_{\text{down}},
\end{equation}
and the computation time consumption is
\begin{equation}
	T^{\text{cmp}}=T^{\text{cmp}}_{\text{train}}+T^{\text{cmp}}_{\text{up}}+ T^{\text{cmp}}_{\text{agg}}+T^{\text{cmp}}_{\text{prep}}+ T^{\text{cmp}}_{\text{pre}}+ T^{\text{cmp}}_{\text{cmit}}+ T^{\text{cmp}}_{\text{rep}}.
\end{equation}

\subsection{Problem Formulation and Analysis}

In this paper, we aim to  minimize the long-term average latency of wireless B-FL by optimizing bandwidth allocation, and transmit power allocation from a long-term perspective.
We assume that the wireless B-FL training process is finished in $\tau$-th training round.
The long-term average latency minimization problem is given by
\begin{subequations}\label{equ_problem}
	\begin{eqnarray}
		\mathop{\text{minimize }}\limits_{\{{\bm{b}^t,\bm{p}^t}\}_{t\in \mathcal{T}} }   &&\frac{1}{\tau} \sum_{t=1}^\tau T({\bm{b}^t,\bm{p}^t}) \nonumber    \\
		\text{subject to }   && \sum_{d\in \mathcal{D} \cup \mathcal{B}} b_{d}^t \le b^{\max}, b_{d}^t\ge 0, \forall t \in \mathcal{T},    \label{equ_constraint_a}  \\
		&& \frac{1}{\tau} \sum_{t=1}^\tau \sum_{d \in \mathcal{B}\cup \mathcal{D}} p_d^t \le \bar{p},  p_d^t \ge 0, \label{equ_constraint_b}
	\end{eqnarray}
\end{subequations}
where  $\mathcal{T}=\{1,2,\dots,\tau\}$ denotes the training round. Here, \eqref{equ_constraint_a}  restricts the maximum system bandwidth and  \eqref{equ_constraint_b} denotes the long-term average power constraint. The long-term average power constraint can provide  dynamic power allocation scheme to  assign more transmit power in worse channel conditions and less power in better channel conditions.

Due to the correlation in the long-term average power constraint, problem \eqref{equ_problem}  is a long-term resource allocation problem, which can not be equivalently transformed to multiple one-shot  problems with respect to the channel of each round. Moreover, the optimization variables of the high-dimensional optimization problem are coupled in the objective function and constraints. As a result, the conventional optimization based algorithms suffer from high computational complexity, which is unacceptable in the time-sensitive  B-FL system \cite{wen2020joint}. Meanwhile, although the supervised learning, and unsupervised learning  based resource allocation schemes can solve the one-shot optimization problem to provide significant reduction on computational complexity,  the sub-optimal performance may occur in the long-term resource allocation problem \cite{shen2021graph}.

In this paper, we proposed a DRL-based algorithm to minimize the long-term cumulative latency by modeling problem (24) as analyze MDP. In particular, the DRL can optimize problem (24) from a long-term perspective  by balancing the latency in current round and that in the future rounds while achieving significant reduction on computational complexity by establishing a direct mapping from the current network information to resource allocation decisions as DNN-based policy.
Besides,  the network information (i.e., channel conditions, current latency) for the decision-making and policy update of DRL agents is obtained by interacting with the dynamic B-FL system, which enables the learned policy to keep track of the environment dynamic.
To this end, we shall formulate the MDP for problem (24), and then we shall develop a TD3 based algorithm to make decisions in continuous action spaces, which will be described in detail in the next section.

\section{TD3 based Resource Allocation Algorithm}
\label{sec_TD3}

In this section, we first formulate the optimization problem as an MDP, and then propose a  TD3 based learning algorithm to solve the MDP problem efficiently.

\subsection{MDP Formulation}

To  obtain  optimal decisions under stochastic environment, we reformulated the optimization problem \eqref{equ_problem} into an MDP, which can be solved by a DRL based scheme. The key components of the MDP include the observed state, action, and reward, which are summarized as follows:

\begin{itemize}
	\item \textit{State}: At each training round, the edge server collects channel state information (CSI) between edge servers and device devices and cumulative  latency, which is necessary for resource allocation. Therefore, The state at time step $t$ is determined by the cumulative  latency and CSI  in the wireless communication system, which is denoted by $s^t$. Particularly, the state contains the channel gain between edge devices and primary edge server, i.e., $\{h^t_{d,B_p} \}, \forall d \in \mathcal{D}$, which has $K$ entries. Besides, the state contains the channel gain between different validator edge servers, i.e., $\{h^t_{d_1,d_2} \}, \forall d_1,d_2 \in \mathcal{B},d_1\neq d_2$, which has $M(M-1)$ entries. Therefore, the state is presented as
	      \begin{equation}
		      \begin{array}{ll}
			      s^{t} = & \left\{\sum_{i=0}^{t-1} T(\bm{b}^i,\bm{p}^i), \{h^t_{d,B_p} \}, \forall d \in \mathcal{D},\right. \\
			              & \qquad \left. \{h^t_{d_1,d_2} \}, \forall d_1,d_2 \in \mathcal{B},d_1\neq d_2 \right\}.
		      \end{array}
	      \end{equation}
	Thus, the dimension of the state space is $K+M(M-1)+1$. \reviseZP{The proposed  DRL-based  resource allocation algorithm can be directly applied when other correlated wireless channel models are considered. However, the  uncorrelated channel (e.g., Rayleigh channel) will decrease the correlation of system state in two consecutive time, which may affect the estimation accuracy of the expected future reward and decrease the performance of resource allocation decisions \cite{wang2022decentralized}.} 

	\item \textit{Action}: From the observed environment states, the agent chooses optimal action, i.e., making decisions based on the policy $\pi$.   The action is constructed by the bandwidth allocation $\bm{b}^t \in \mathbb{R}^{M+K}$ and transmit power allocation $\bm{p}^t \in \mathbb{R}^{M+K}$ of all edge servers and edge devices at training round $t$, i.e.,
	      \begin{equation}
		      a^{t}=\left\{ \bm{b}^t,\bm{p}^t     \right\}.
	      \end{equation}
	Therefore,  the dimension of the action space is $2(M+K)$. 

	\item \textit{Reward}: After the action $a^{t}$ has been taken by the environment, the DRL agent will obtain a reward from the environment and update the policy $\pi$ to maximize the reward in the future. To minimize the latency in wireless B-FL training process, the reward at the $t$-th step is defined by negative latency, i.e., maximum reward means minimum latency, which is presented  as
	      \begin{equation}
		      r^t=\left\{ \begin{array}{ll}
			      -T(\bm{b}^t,\bm{p}^t) & \text{when \eqref{equ_constraint_a}, \eqref{equ_constraint_b} are satisfied,} \\
			      r_p                   & \text{otherwise,}                                                             \\
		      \end{array}
		      \right.
	      \end{equation}
	      When the constraints \eqref{equ_constraint_a} and \eqref{equ_constraint_b} are satisfied, the immediate reward is the negative of the latency in this round. When the constraints \eqref{equ_constraint_a} and \eqref{equ_constraint_b} are not  satisfied, the immediate reward is set to $r_p$. $r_p$ is an extremely small value, which is regarded as a penalty. Therefore,  the actions selected by the agent will satisfy the constraints \eqref{equ_constraint_a} and
	      \eqref{equ_constraint_b}.

	\item \textit{MDP Optimization Problem}:
	      The action-value function is defined as
	      \begin{equation}
		      Q_{\pi_\phi}(s^t,a^t)=\mathbb{E}_{\pi_\phi} \left[ \sum_{i=0}^{\infty} \gamma^i r^{t+i} |s^t,a^t  \right],
	      \end{equation}
	      where $  \sum_{i=0}^{\infty} \gamma^i r^{t+i} $ denotes the total discount reward from the $i$-th training round and $\gamma\in[0,1]$ represents the discount factor.
	      Therefore, the action-value function expresses a ``myopic'' evaluation if $\gamma \rightarrow 0$ or ``fast-sighted'' evaluation if $\gamma \rightarrow 1$. The action-value function can evaluate the action selected by a policy $\pi_\phi$ with parameters $\phi$.
	      Moreover, the action-value function can be rewritten as the Bellman equation, i.e.,
	      \begin{equation}
		      Q_{\pi_\phi}(s^t,a^t)=\mathbb{E}_{\pi_\phi} \left[ r^t+\gamma Q_{\pi_\phi}(s^{(t+1)},a^{(t+1)}) | s^t,a^t  \right].
	      \end{equation}
	      Therefore, optimal policy can be obtained by maximizing the action-value function to solve the MDP problem, i.e.,
	      \begin{equation}
		      \pi_*=\arg \min_{\pi_{\phi}} Q_{\pi_\phi}(s^t,a^t).
	      \end{equation}

\end{itemize}

\reviseZP{The resource allocation problem \eqref{equ_problem} is reformulated as an MDP without having state transition probabilities $P(s^{t+1}|s^t,a^t)$  in prior due to the complexity and dynamics of wireless B-FL systems.  Therefore, we resort to adopting   model-free  reinforcement learning based on value function approximation, which does not require the transition probability distribution to make decisions.}

\begin{algorithm}[t]
	\caption{TD3-based bandwidth and transmit power allocation algorithm}
	\label{alg_TD3}
	\begin{algorithmic}[1]
		\Ensure {Actor network $\pi_\phi$ and critic networks $\{Q_{\theta_1},Q_{\theta_2}\}$}

		\State {Initialize critic networks $\{Q_{\theta_1},Q_{\theta_2}\}$and actor network $\pi_\phi$ with random parameters $\theta_1,\theta_2,\pi_\phi$.}
		\State {Initialize target networks $\theta_1' \leftarrow \theta_1,\theta_2' \leftarrow \theta_2, \phi'\leftarrow \phi$.}
		\State {Initialize replay buffer $\mathcal{R}$.}
		\State {Initialize noise range $c$, noise variance $\sigma_1^2,\sigma_2^2$, update proportion $\delta$, discount factor $\gamma$.}
		\State {Explore $E$ time steps with random policy, obtain the reward $r^t$ and the next state $s^{t+1}$, and storage the transition $(s^t, a^t, r^t, s^{t+1})$ of the $E$ steps into $\mathcal{R}$.}
		\While{$\text{time step} \le P$}
		\State {Select action with exploration noise \Statex \qquad $a^t=\pi_\phi(s^t) +n_1,n_1 \sim \mathcal{N}(0,\sigma^2_1)$. }
		\State {Obtain bandwidth allocation $\bm{b}^t$ and transmit power \Statex \qquad allocation $\bm{p}^t$ from action $a^t$, and assign them to edge \Statex \qquad servers  and device devices. }
		\State {Edge devices and edge servers begin B-FL  training \Statex \qquad process and  get the latency in one round.}
		\State {Calculate reward $r^t$, update state $s^{t+1}$, and storage the \Statex \qquad transition $(s^t, a^t, r^t, s^{t+1})$ into $\mathcal{R}$.}
		\State {Sample a mini-batch of transitions from $\mathcal{R}$.}
		\State {Obtain target action $a'^{t+1}=\pi_{\phi'}(s^{t+1}) + n_2, n_2 $\Statex \qquad $\sim \text{clip}(\mathcal{N}(0,\sigma_2^2),-c,c)$.}
		\State {$y=r^t+\gamma \min_{i=1,2} Q_{\theta_i'}(s^{t+1},a'^{t+1})$.}
		\State {Update critics $\theta_i \leftarrow \arg\min_{\theta_i}\mathbb{E} \left[ (y- Q_{\theta_i}(s^{t},a^{t})   )^2   \right] $.}
		\State {Update $\phi$ using the deterministic policy gradient\Statex \qquad  algorithm.}
		\If{step mod $\vartheta =0$}
		\State {Update the target networks:}
		\State {$	\theta_1' =  \kappa \theta_1 +(1-\kappa)\theta_1'$,}
		\State {$\theta_2' =  \kappa \theta_2 +(1-\kappa)\theta_2'\nonumber$,}
		\State {$\phi' = \kappa \phi+(1-\kappa)\phi'$.}
		\EndIf
		\State {$t \leftarrow t+1$}
		\EndWhile
	\end{algorithmic}
\end{algorithm}

\subsection{TD3 algorithm}\label{sec_TD3_TD3}
\begin{figure*}[t]
	\centering
	\includegraphics[width=1\linewidth]{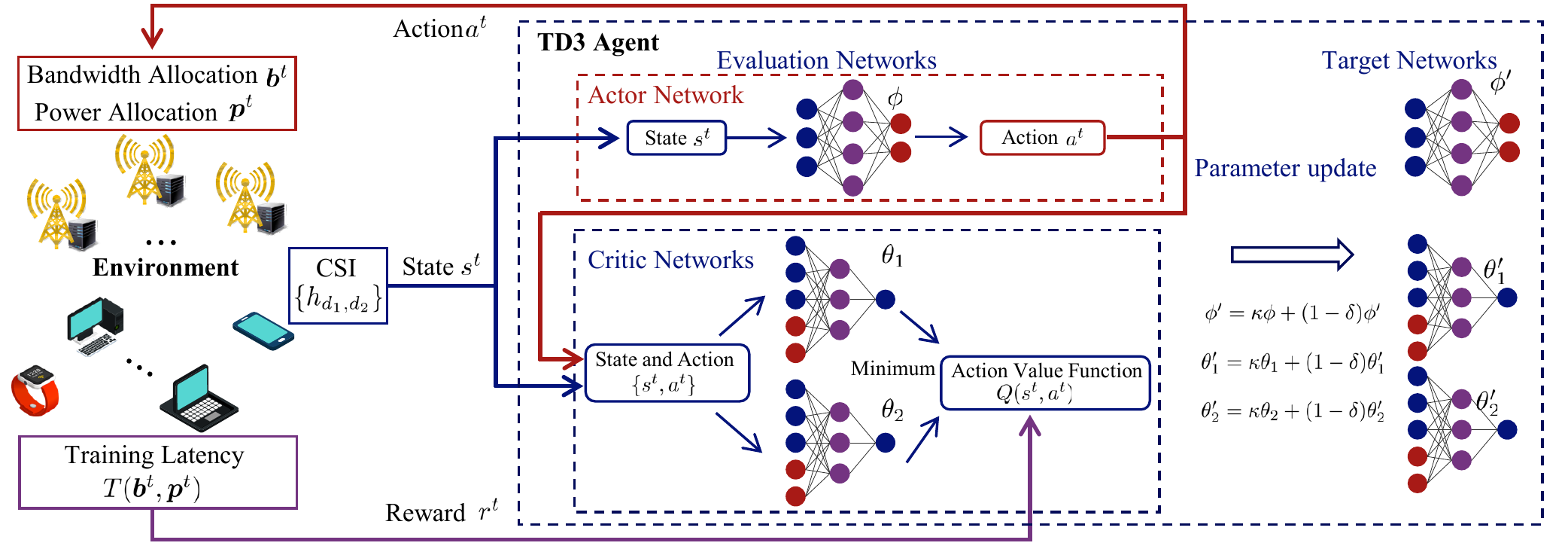} 
	\caption{Framework of TD3 algorithm.}\label{fig_TD3}
\end{figure*}

\begin{figure}
	\centering
	\subfigure[Structure of critic network.]{\label{fig_c_structures}
		\begin{minipage}[b]{0.4\textwidth}
			\centering
			\includegraphics[width=1\textwidth]{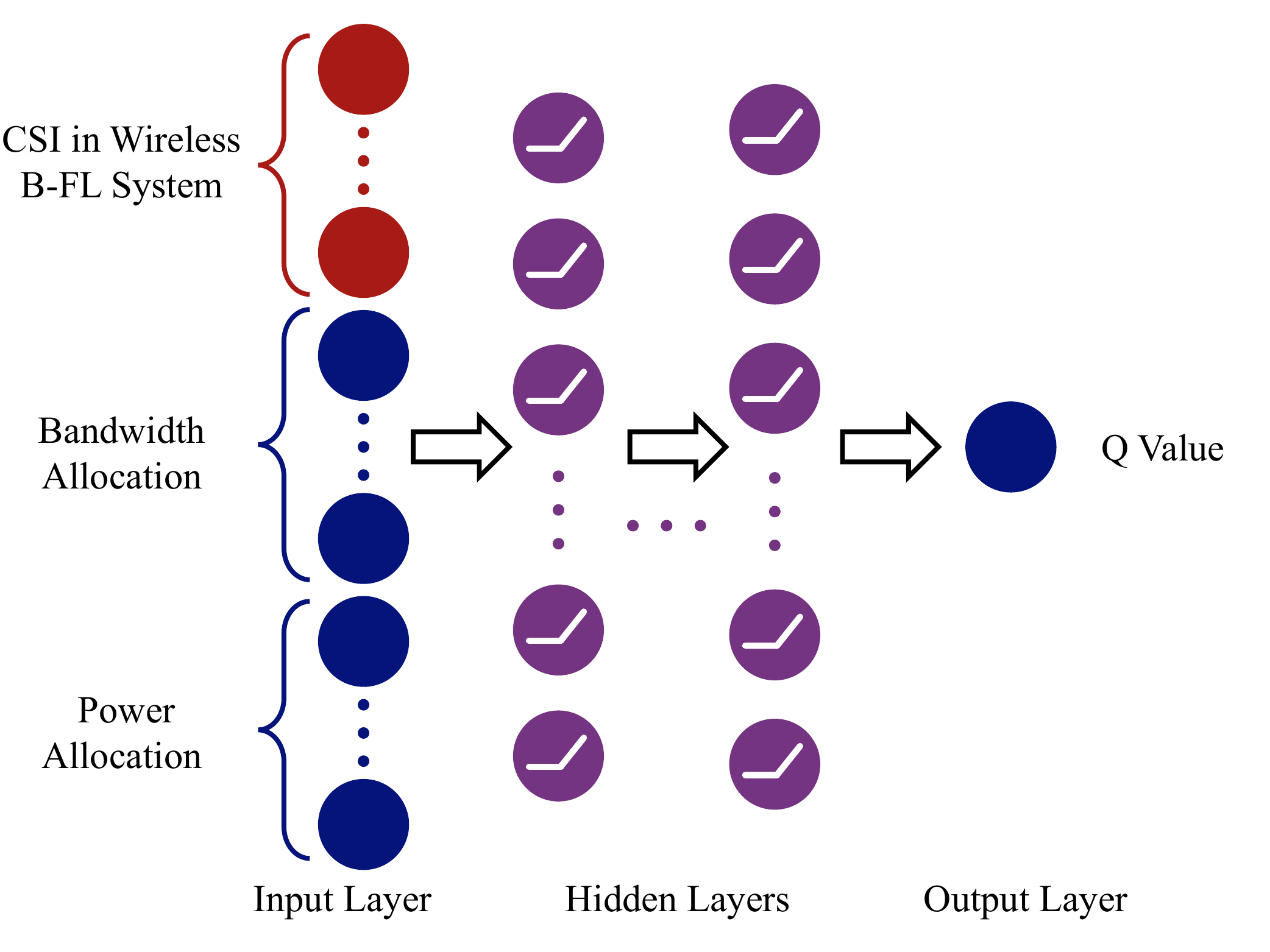}
		\end{minipage}
	}
	\subfigure[Structure of actor network.]{\label{fig_a_structures}
		\begin{minipage}[b]{0.4\textwidth}
			\centering
			\includegraphics[width=1\textwidth]{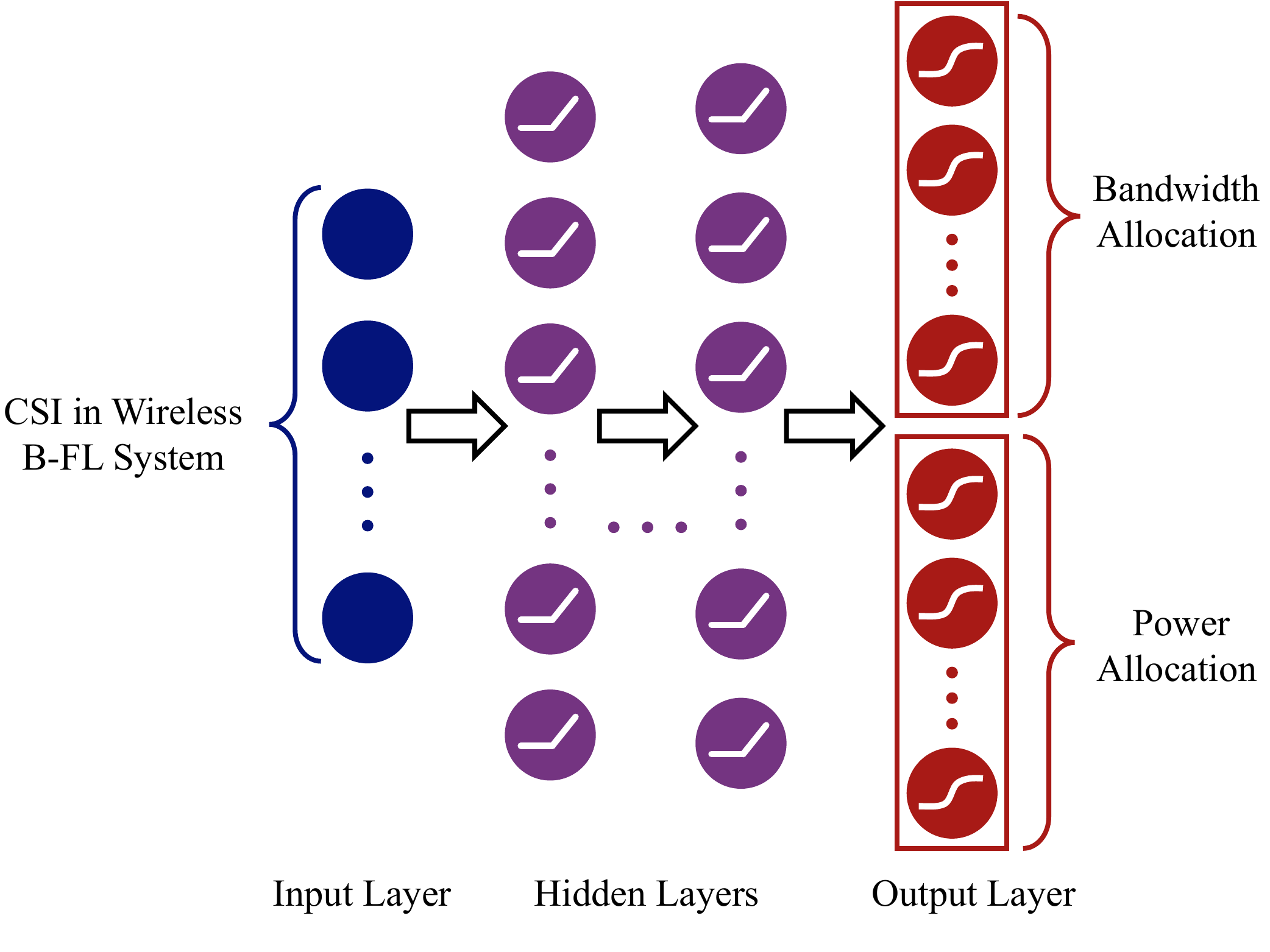}
		\end{minipage}
	}
	\caption{Proposed structures of actor and critic networks in TD3 algorithm.  }
	\label{fig_structures}
\end{figure}

From the definition of the MDP problem, the state space and action space are continuous, which indicates that infinite dimension of action  can not be solved by Q-learning or deep Q network. Recently, deep deterministic policy gradient (DDPG) algorithm has achieved considerable performance for solving MDP problem with continuous action variables. Moreover, the enhanced version of DDPG, i.e., twin delayed deep deterministic policy gradient (TD3) \cite{fujimoto2018addressing}, can reduce the overestimation bias of Q value and avoid sub-optimal policy updates,
which improves the policy learning speed and performance in continuous control domains.
The TD3 is widely used to solve complex resource allocation  problem in real-time.
We thus propose a TD3 based resource allocation algorithm to solve the MDP problem of B-FL system in this paper.
The TD3 algorithm is presented in Algorithm \ref{alg_TD3} in details.

Particularly, the actor-critic architecture is adopted in TD3 algorithm, which combines the value-based and policy-based reinforcement learning algorithm and is shown in Fig. \ref{fig_TD3}. In the actor-critic architecture, the actor will output the selected actions in continuous space from the input observed state, i.e., the policy $a^t=\pi_\phi(s^t)$.
The critic is used to approximate the action-value function, which outputs Q value from the inputs selected action and current state, i.e., $Q(s^t,a^t)$.
Due to the complexity of the mapping function of actor and critic, and vector structure of state and action, the actor and critic are implemented by deep neural networks (DNN), which consist of several full-connected layers of neurons to  construct the feature extractor and the output layer. Then, we introduce the training processes of  actor-critic architecture and update rules of TD3 algorithm as follows.

\subsubsection{Critic Network Training}
The critic network  aims to estimate the real Q table in continuous state and action spaces as shown in Fig. \ref{fig_c_structures}, which approximates the  mapping function from state and action values to Q value, i.e., $Q_{\theta}(s^t,a^t)$. Therefore, the loss function of critic network is the mean square error (MSE) between estimated value and the target value, i.e.,
\begin{equation}\label{equ_critic_loss}
	L(\theta) = \mathbb{E} \left[\left( y- Q_{\theta}(s^t,a^t) \right)^2 \right],
\end{equation}
where $y$ is the target value and $Q_{\theta}(s^t,a^t)$ is the estimated value under DNN parameter $\theta$. Especially, $y$ is generated by target critic network, i.e.,
\begin{equation}
	y=r^t+\gamma Q_{\theta'}(s^{t+1},a'^{t+1}),
\end{equation}
where $\theta'$ is DNN parameter in target network of critic network and $\gamma $ is discount factor. $\theta'$ is the periodically updated parameters of the target critic $Q_{\theta'}(s^{t},a'^{t})$, and $a'^{t+1}$ is the action taken in the next state $s^{t+1}$ selected by the target actor $\pi_{\phi'}$.

However, the update of the value function with the above target value have an overestimate bias, which will also affect the accurate update of the actor \cite{fujimoto2018addressing}. To address the bias problem, the TD3 uses two approximately independent target critics $\left\{Q_{\theta_1}, Q_{\theta_2} \right\}$ to estimate the value function, in which $Q_{\theta_1}$ is a biased estimator and $Q_{\theta_2} $ is an estimator with less overestimate bias. The minimum of the two estimators is used in the update of the value function, i.e.,
\begin{equation}
	y=r^t+\gamma \min \left\{Q_{\theta_1'}(s^{t+1},a'^{t+1}), Q_{\theta_2'}(s^{t+1},a'^{t+1}) \right\}.
\end{equation}
Under the loss function in \eqref{equ_critic_loss}, the update rules of the two critic network can be written as
\begin{equation}
	\theta_i = \theta_i - \eta_c 	\nabla_{\theta_i} L(\theta_i),\forall i=1,2,
\end{equation}
where $\eta_c$ is learning rate of critic network, and $\nabla_{\theta_i} L(\theta_i)$ is the gradient of $i$-th  critic network, i.e.,
\begin{equation}
	\nabla_{\theta_i} L(\theta_i)= - \mathbb{E}_{s^t} \left[\left( y_j-Q_i(s_j,a_j|\theta_i)\right) \nabla_{\theta_i} Q_i(s_j,a_j|\theta_i) \right].
\end{equation}

\subsubsection{Actor Network Training}
The actor network  aims to select optimal actions in continuous state and action spaces as shown in Fig. \ref{fig_a_structures}, which approximates the  mapping function from state to action, i.e., $a^t=\pi_\phi(s^t)$. \reviseZP{The output layer of the actor network is designed to have neurons for bandwidth allocation and power allocation.  The first  set of output neurons  is implemented by softmax function, which  guarantees the sum of output is equal to 1 to  represent the  percentage of bandwidth allocation.   The second set of output neurons  is implemented by Sigmoid function, which guarantees each output is between 0 and 1 to represent the percentage of the maximum transmit power.}  The optimal action is selected to achieve the maximum $\mathbb{E} \left[ Q_1(s^t,a^t|\theta_1) \right]$, which utilizes the DNN parameter of the first critic network to update actor network.
Accordingly, the actor can be updated using the deterministic policy gradient algorithm (DPG), i.e.,
\begin{equation}
	\nabla_\phi J(\phi)=\mathbb{E}_{\pi_\phi} \nabla_a \left[   Q_{\pi_\phi}(s^{t+1},a^{t+1}) |_{a=\pi_\phi(s^t)} \nabla_\phi \pi_\phi(s^t)   \right],
\end{equation}
where $\pi_\phi(s^t)$ is actor network with DNN parameter $\phi$. Therefore, the update rule of actor network is
\begin{equation}\label{equ_actor_update}
	\phi = \phi - \eta_a 	\nabla_\phi J(\phi),
\end{equation}
where $\eta_a$ is learning rate of actor network.

\subsubsection{Target Networks Updating}

The two critic networks and one actor network have corresponding target networks, which are called target critic networks and target actor network.
The target critic networks aim to generate target values, which approximates real Q values.
The target actor network is used to select actions.
In other words, the critic networks and actor network, a.k.a., online networks, are used to update the DNN parameters in each step, and the target networks are assumed as real Q value and optimal action. The DNN parameters of target networks are updated based on the online networks with delayed method, i.e., the target networks update after several steps of online networks.
In general, the update rule of target networks is
\begin{eqnarray}
	\theta_1' =  \kappa \theta_1 +(1-\kappa)\theta_1' \label{equ_target_update_1}, \\
	\theta_2' =  \kappa \theta_2 +(1-\kappa)\theta_2',\\
	\phi' = \kappa \phi+(1-\kappa)\phi'  \label{equ_target_update_3},
\end{eqnarray}
where $\kappa$ is update proportion, i.e., the proportion of DNN parameters of online networks.

In summary, there are two types of networks, i.e., two critic networks and one actor network. Each of them contains two sub-nets, i.e., online network and target network. These six DNNs consist various  layers and all layers contain their corresponding parameters. Actor network parameters $\{\phi,\phi' \}$  denote online network parameter and target network parameter respectively. Critic network parameters $\{\theta_1,\theta_2,\theta_1',\theta_2' \}$  indicate online network parameter and target network parameter  of the two critic networks, respectively.

Besides, the TD3 uses a reply memory buffer $\mathcal{R}$ to store the experience $(s^t,a^t,r^t,s^{t+1})$ from the interaction between agent and environment. In the training process, a mini-batch of experience is randomly selected from the reply memory buffer $\mathcal{R}$ to update the parameters of the online networks. Particularly, the online critic networks are updated by minimizing the loss function \eqref{equ_critic_loss} and the online actor network is updated base on \eqref{equ_actor_update}. After every $\vartheta$ steps, the parameters of targets networks are updated \eqref{equ_target_update_1}-\eqref{equ_target_update_3} with update proportion $\delta$. The pseudocode of TD3 based resource allocation algorithm is presented in Algorithm \ref{alg_TD3}.

\reviseZP{
	We assume the number of edge servers and edge devices is invariable during the FL training process. The B-FL  can still be applied when the edge servers or edge devices exit the system before the completion of FL, but suffer from system security and global model accuracy degradation. 
    The DRL-based  resource allocation algorithm is compatible with the case of moving edge servers or edge devices. When the edge servers or edge devices exit the system, the relevant parameters in the  states and actions are set to be null to cope with the smaller number of edge servers or edge devices \cite{guo2020adaptive}. However, if the number of edge servers and edge devices increases, the DRL-based resource allocation algorithm needs to be retrained, because the output layer dimension of actor network increases.
  } 

\subsection{\reviseZP{Complexity Analysis}}

  \reviseZP{
	Based on the description of  TD3 algorithm, the training process of the DRL for the proposed B-FL systems  can be performed offline with simulated channel states to train the actor network and critic network. Then, the trained actor network models are deployed online to allocate the wireless resources based on the system state to minimize the long-term latency of B-FL. Furthermore, the experience $(s^t, a^t, r^t, s^{t+1})$ generated by online actor network can be collected, which is used to train and update the actor network.  }

     \reviseZP{
		In the training phase, 2 actor networks and 4 critic networks need to be trained, the most significant complexity is caused by the back propagation. Hence, the computational complexity is $\mathcal{O}\left( 2 \sum_{l=1}^{L^a}z^a_l z^a_{l+1}+4\sum_{l=1}^{L^c}z^c_l z^c_{l+1}\right)$, where $L^c$ is the number of layers in critic network. Let $z^c_l $ denotes the number of neurons in $l$-th layer of critic network. Hence, $z^c_1=K+M(M-1)+1+2(M+K)$ is the number of neurons in input layer and $z^a_{L}=1$ is the number of neurons in  output layer. Since the proposed algorithm selected  $Q$ mini-batch transitions to  train the actor networks and critic networks and takes $P$ training step  in total. Therefore, total computational complexity is $\mathcal{O}\left( P Q  \left(  2\sum_{l=1}^{L^a}z^a_l z^a_{l+1}+4\sum_{l=1}^{L^c}z^c_l z^c_{l+1}\right) \right)$.
          }

    \reviseZP{
		In the deployment phase,  for each single decision-making, the complexity only caused by the actor network, which can be calculated as $\mathcal{O}\left( \sum_{l=1}^{L^a}z^a_l z^a_{l+1}\right)$, Let $L^a$ is the number of layers in actor network. The $z^a_l $ denotes the number of neurons in $l$-th layer of actor network. Hence, $z^a_1=K+M(M-1)+1$ is the number of neurons in  input layer and $z^a_{L}=2(M+K)$ is the number of neurons in  output layer.}

\section{Numerical Results and Analysis} \label{sec_results}

In this section, various simulation results are presented to show the performance of wireless B-FL and related TD3 based resource allocation algorithm. We first introduce the simulation settings, and then show the results of proposed algorithms.

\begin{table*}\footnotesize
	\newcommand{\tabincell}[2]{\begin{tabular}{@{}#1@{}}#2.8\end{tabular}}
	\renewcommand\arraystretch{1.2}
	\caption{
		\vspace*{-0.1em}The Accuracies of FL and B-FL with MINST Dataset over Different Percentage of Malicious Devices.}\vspace*{-0.6em}
	\centering  \label{tab_acc}
	\begin{tabular}{|c|c|c|c|c|c|c|c|c|c|c|c|}
		\hline
		Malicious edge devices percentage & 0\%   & 10\%  & 20\%  & 30\%  & 40\%  & 50\%  & 60\%  & 70\%  & 80\%  & 90\%  & 100\% \\
		\hline
		FL with FedAvg                    & 97.92 & 95.59 & 92.38 & 89.67 & 86.42 & 11.35 & 11.35 & 11.35 & 11.35 & 10.28 & 9.74  \\
		\hline
		B-FL with multi-KRUM              & 97.93 & 97.63 & 97.93 & 97.80 & 97.90 & 93.74 & 88.14 & 11.35 & 11.35 & 9.80  & 9.74  \\
		\hline
	\end{tabular}
	\vspace{-0.3cm}
\end{table*}

\subsection{Settings and Benchmarks}
In the following simulations, we consider a wireless B-FL system consisting of several edge servers and edge devices to execute ML task, which guarantees  security and privacy in the training process based on blockchain technique. The details of the system model and simulation scenario are defined as follows.

\subsubsection{System Settings}
In our simulation, we consider a circular wireless network with $M=4$ edge server serving for $K=10$ edge devices, and the location of each edge server or edge device is uniformly distributed  in a circle with a radius of $100$ meters. The duration of time-slot $T_0$ is set as $10$ms according to LTE standard and Doppler frequency $f_d$ is $5$Hz due to the assumption of low mobility in B-FL system.  The path loss exponent $\alpha$ is $2.5$.
In general, we assume that the maximum CPU frequencies of edge servers and devices are $2.4$GHz and $1$GHz, respectively. The maximum system transmission powers are set as $24$dBm and the maximum system bandwidth is $100 $MHz. The noise power spectral density is  $N_0=-174$dBm/Hz.

\subsubsection{MINST Dataset for Handwriting Recognition}

\begin{figure}[t]
	\centering
	\includegraphics[width=1\linewidth]{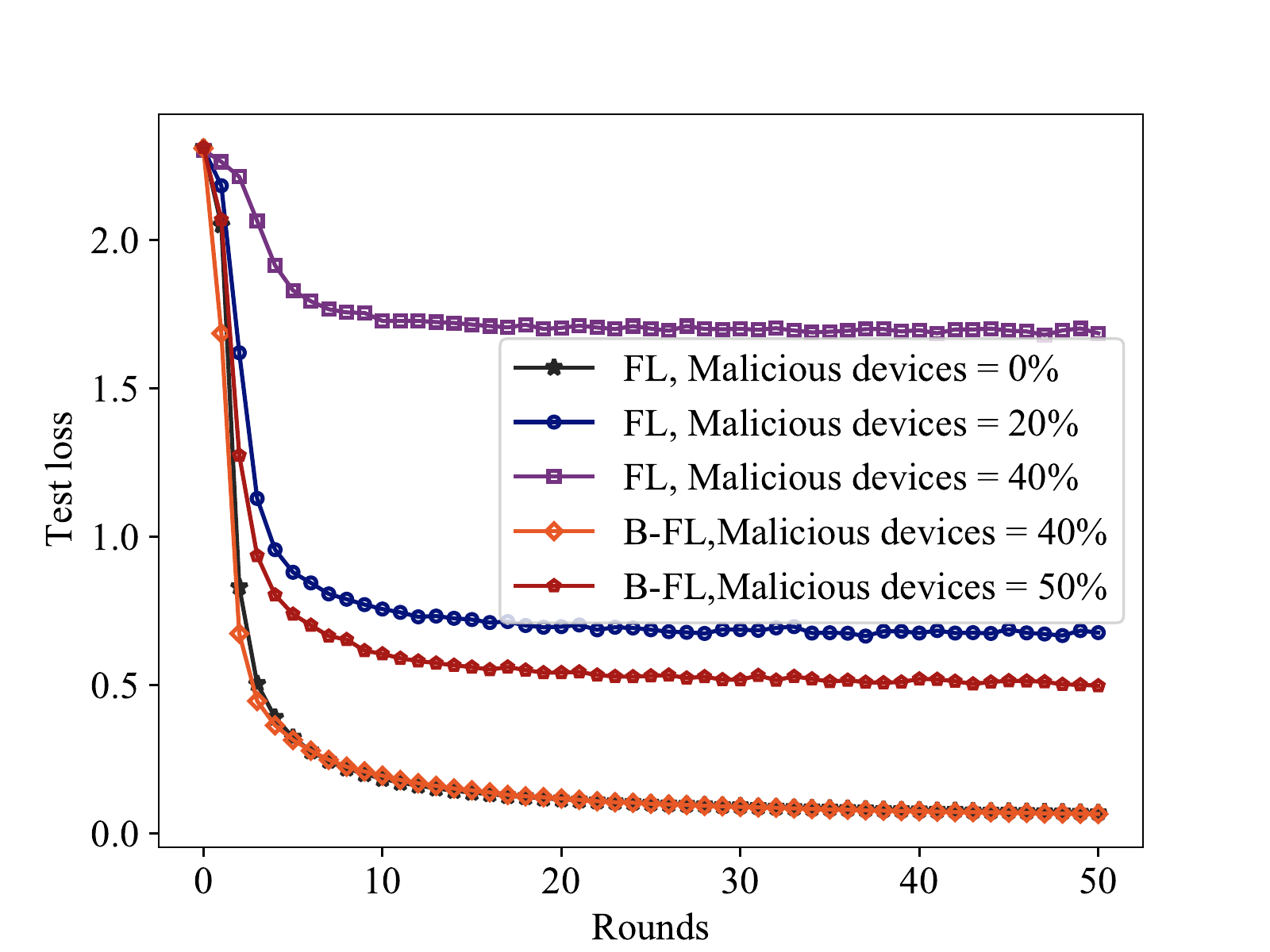} 
	\vspace{-4mm}
	\caption{Test loss of FL and B-FL with malicious devices for handwriting recognition. }\label{fig_loss_minist}
	\vspace{-4mm}
\end{figure}
\begin{figure}[t]
	\centering
	\includegraphics[width=1\linewidth]{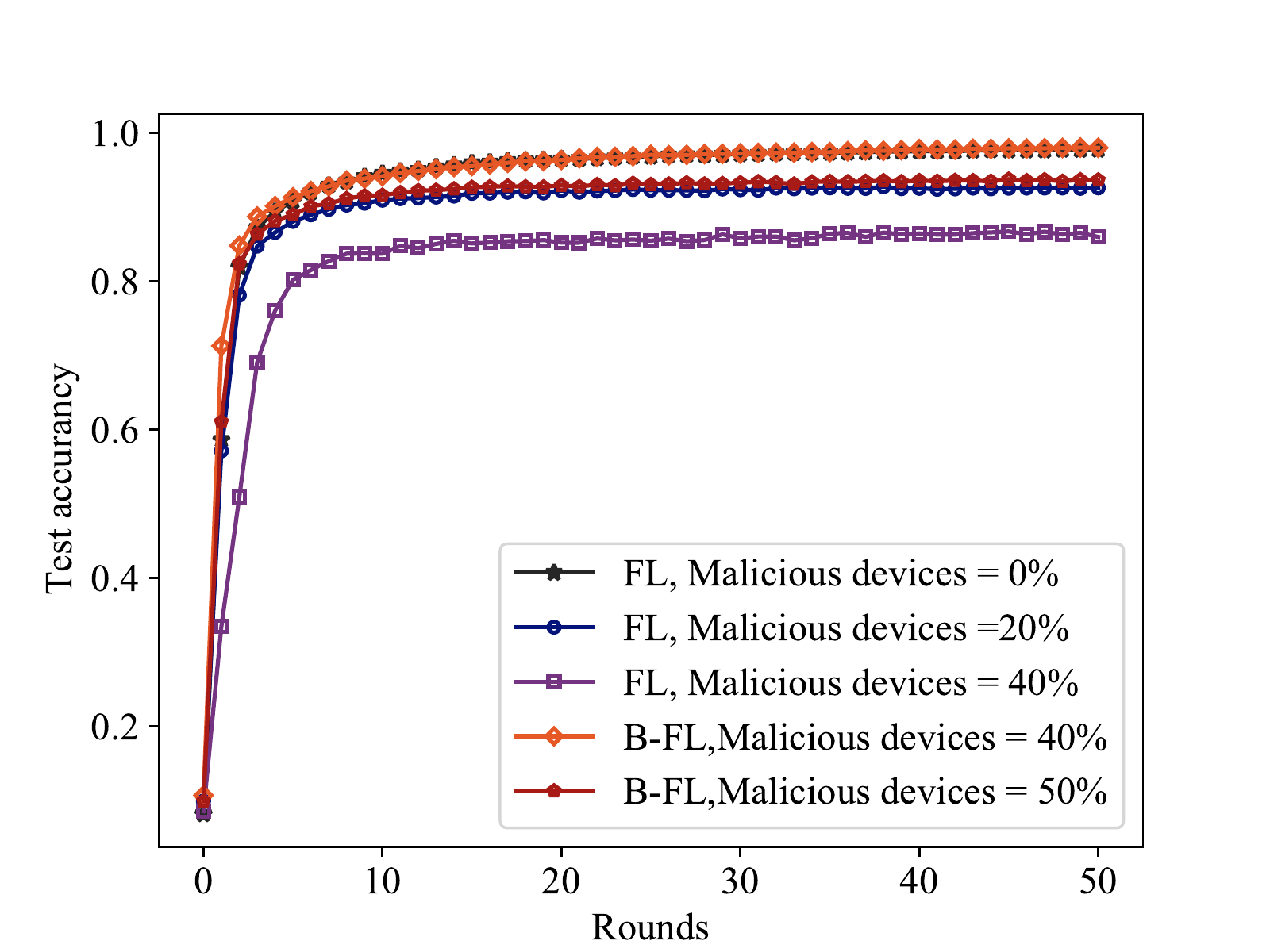} 
	\vspace{-4mm}
	\caption{Test accuracy of FL and B-FL with malicious edge devices for handwriting recognition.}\label{fig_accuracy_minist}
	\vspace{-4mm}
\end{figure}

To simulate the performance of FL and B-FL, we consider training a CNN model based on MINST dataset, which is a standard dataset used for testing ML algorithm and consists of $60,000$ training handwritten image  samples and $10,000$ test handwritten image samples. The handwritten image samples are $28\times 28$ gray scale images denoting a handwritten number within $0$ to $9$. The dataset  gives a typical and proper complexity in ML tasks. We design the convolutional neural network (CNN) model by two $5\times 5$ convolutional layers (10 channels and 20 channels respectively), two $2\times 2$ max-pooling layers, two fully connected layers,  and two dropout layers. The rectified linear unit (ReLU) activation is used in each layer and the softmax activation is employed in output layer. Therefore, each device randomly has 6000 training samples and 1000 test samples. In the training process, each device executes 2 local training epoch with 128 batch size and the  learning rate is set as $0.01$.

\reviseZP{\subsubsection{CIFAR-10 Dataset for Image Classification}  To verify the performance of B-FL in practical real-word dataset, we utilize more popular and standard CIFAR-10  dataset and train the AlexNet model. The CIFAR-10  consists of 50000 training  $32\times32$ color images in 10 classes and 10000 test images. The AlexNet is a typical CNN model with  5 convolutional layers and max pooling  and 3 fully-connected layers.  Therefore, each device randomly has 5000 training samples and executes 1 local training epoch with 128 batch size and  the learning rate is set as 0.01.}

\subsubsection{Heart Activity Dataset for Affect Recognition}

More and more researches have shown great interest to improve the health of individuals via prevention and diagnostics by using sensors of Internet of Medical Thing (IoMT). In particular,  affect recognition of stress is remarkable to improve the health of individuals based on the biomedical informatics collected from wearable IoT devices. In this paper, we utilize the preprocessed dataset of 26 individuals from the paper \cite{can2021privacypreserving}. The dataset  consists of heart activity samples range from 60 to 125, which means a non-independent identically distributed (non-iid) dataset. Each sample has a $16$ dimension feature vector   and a label to indicate low stress and high stress, which means it is a 2-class stress-level classification.  We design the fully-connected neural network (FNN) model by two hidden layers with ReLU activation and  one output layer with sigmoid activation. Each hidden layer is composed of 100 neurons. In the training process, we set that the batch size is 32, the local learning rate is 5e-6 and local training epoch is 1. Furthermore, we partition the 26 individuals as  20 training edge devices to train a shared global model and 6 test edge devices to evaluate the accuracy of global model.

\subsubsection{TD3 Algorithm Settings}

Base on the details of TD3 algorithm in Section \ref{sec_TD3}, we implement the proposed algorithm by TensorFlow in a general computer with CPU Intel Xeon E5-2643 v4 and GPU Nvidia GeForce GTX 1080 Ti. The actor network is designed by 5 hidden layers, which has 512, 1024, 2048, 1024, and 512 neurons with ReLU activation respectively. The output layer is a softmax layer to output action with constrains. The critic network is constructed by 4 hidden layers, which has 512, 1024, 512, and 512 neurons with ReLU activation respectively. The output layer is a linear layer to output Q value.
The hyper-parameters in TD3 algorithm are set as follows, i.e.,  the size of replay buffer is $10^6$,  update proportion $\kappa=5\times 10^{-3}$, discount factor $\gamma=0.99$, steps of exploration $E = 512$, update frequency $ \vartheta= 2$, and the learning rate of the actor and critic networks is $\eta_a=\eta_c=1\times 10^{-4}$. We train the networks with max steps $5000$, which can achieve stable performance in our experiments.

\subsubsection{Benchmarks}
To verify the effectiveness of our proposed TD3-based resource allocation algorithm, we choose three other algorithms to compare, i.e.,
\begin{itemize}
	\item \textit{Random Allocation}: The random allocation scheme  allocates the bandwidth and transmit power to edge servers and devices from uniformly random distribution.

	\item \textit{Average Allocation}: The average allocation scheme uniformly allocates the bandwidth and transmit power to edge servers and devices, i.e., all edge servers and edge devices are assigned with the same resources.

	\item \textit{Monte Carlo Algorithm}: The Monte Carlo algorithm randomly samples $C$ allocation solutions and choose the best one, i.e., the latency is the smallest. Therefore, the best allocation solution is close to the global optimal solution if the $C$ is large enough. We set $C=10^6$ in the paper.
\end{itemize}

\begin{figure}[t]
	\centering
	\includegraphics[width=1\linewidth]{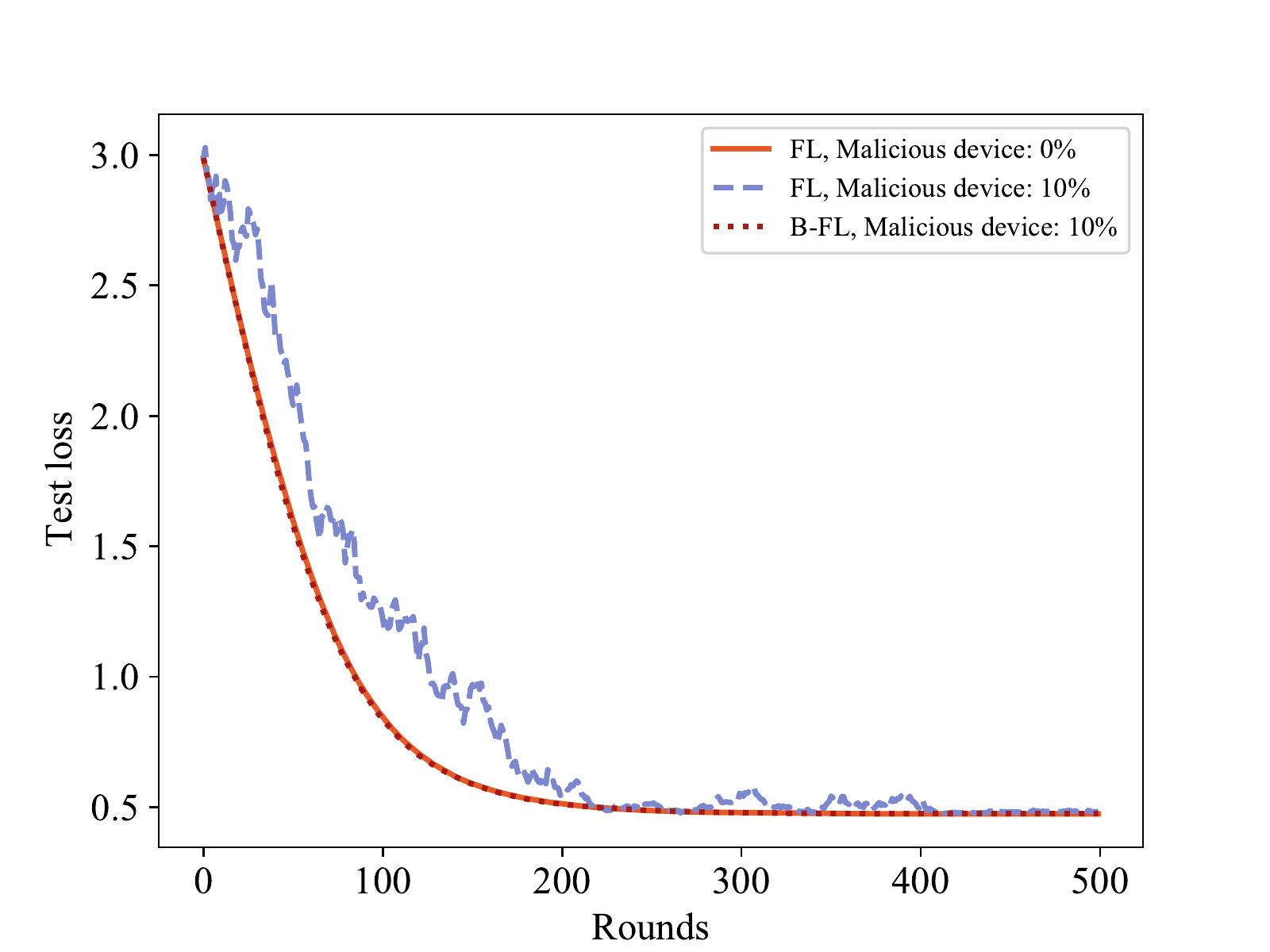} 
	\vspace{-4mm}
	\caption{Test loss of FL and B-FL with malicious devices for affect recognition. }\label{fig_accurancy_fig}
	\vspace{-4mm}
\end{figure}
\begin{figure}[t]
	\centering
	\includegraphics[width=1\linewidth]{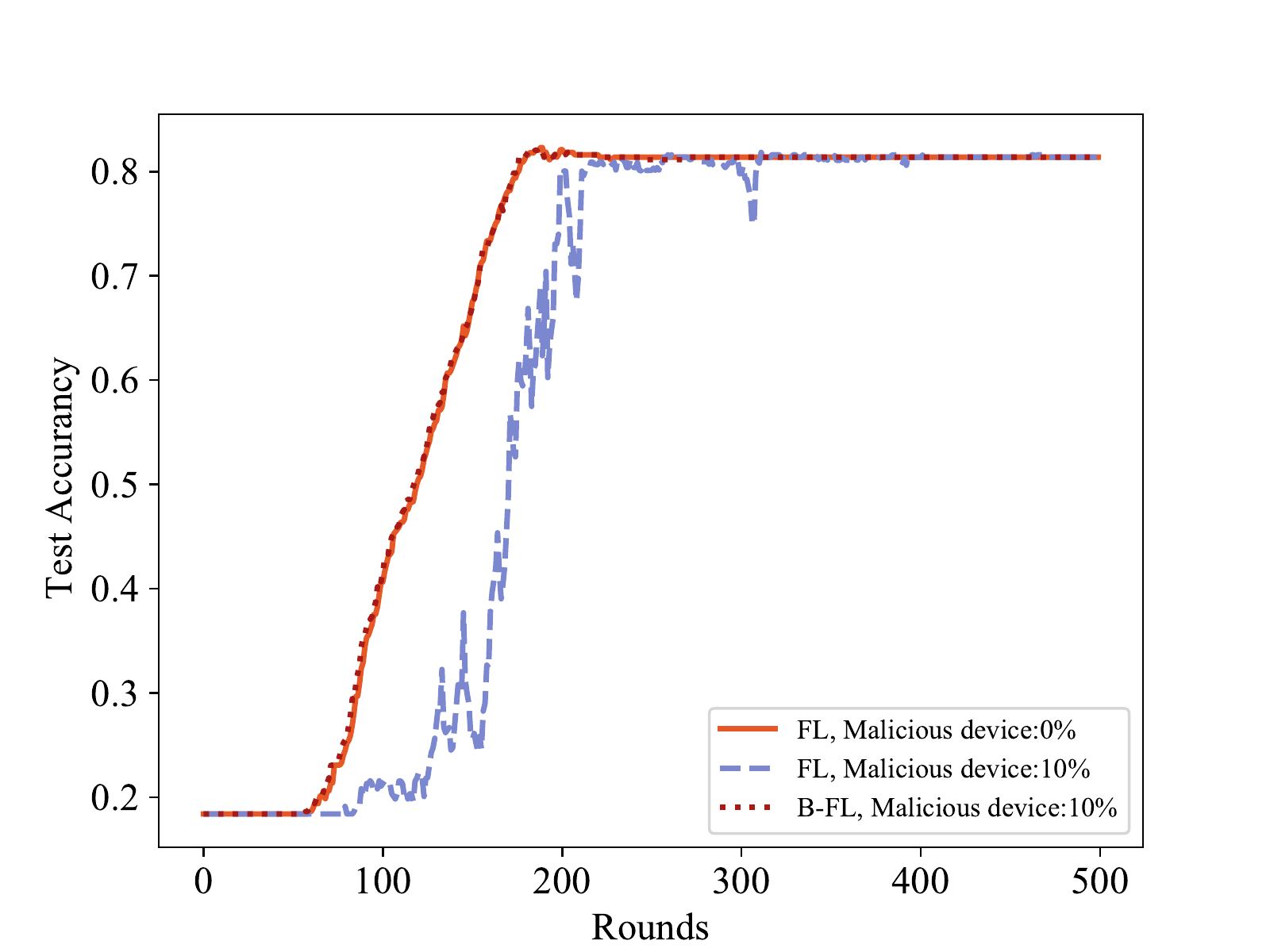} 
	\vspace{-4mm}
	\caption{Test accuracy of FL and B-FL with malicious edge devices for affect recognition.}\label{fig_loss_fig}
	\vspace{-4mm}
\end{figure}

\begin{figure}[htbp]
	\centering
	\begin{minipage}[t]{0.48\textwidth}
		\centering
		\includegraphics[width=1\textwidth]{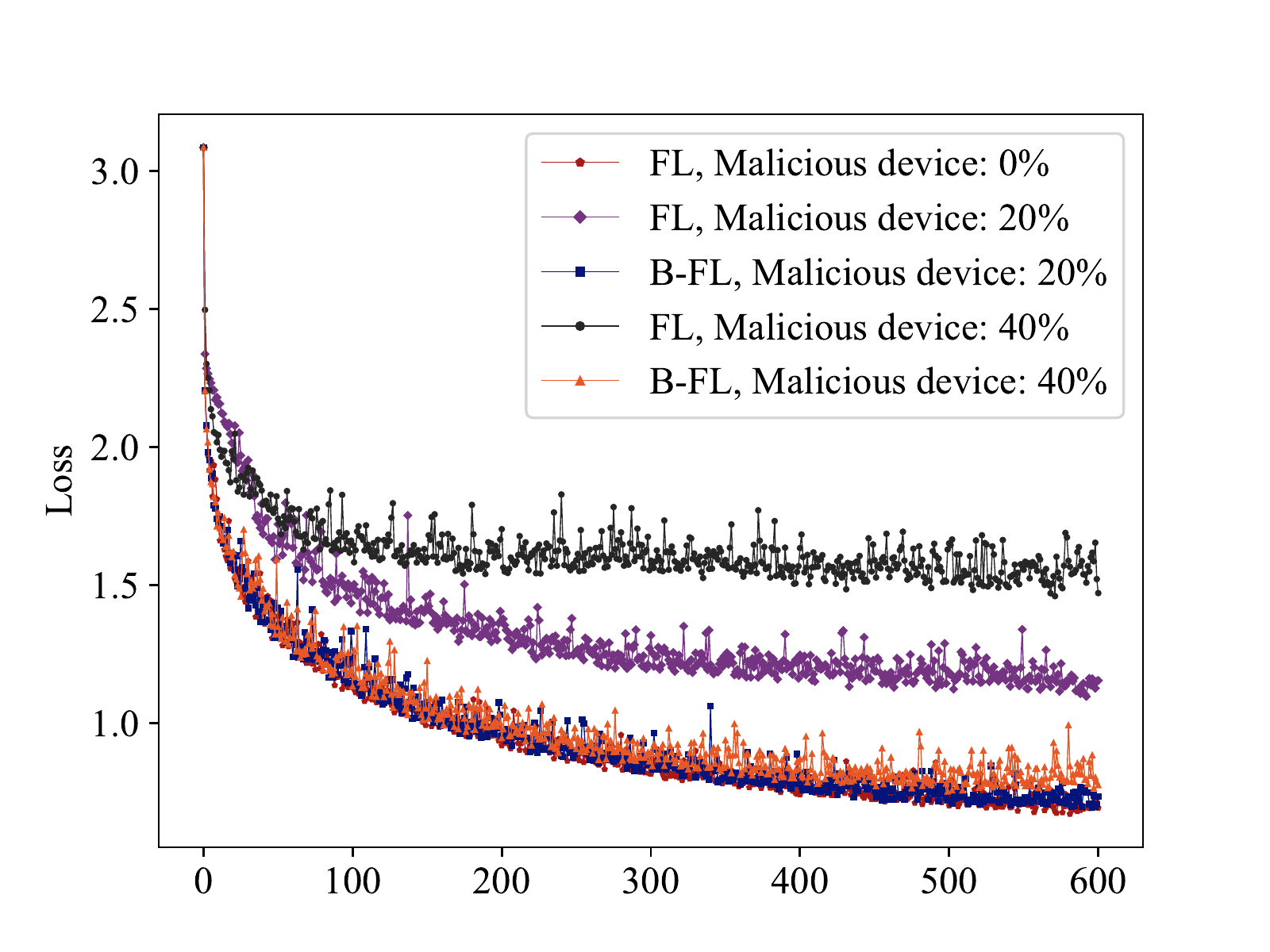}
		\caption{Test loss of FL and B-FL with malicious edge devices for CIFAR10 dataset.}\label{fig_loss_cifar10}
	\end{minipage}
	\begin{minipage}[t]{0.48\textwidth}
		\centering
		\includegraphics[width=1\textwidth]{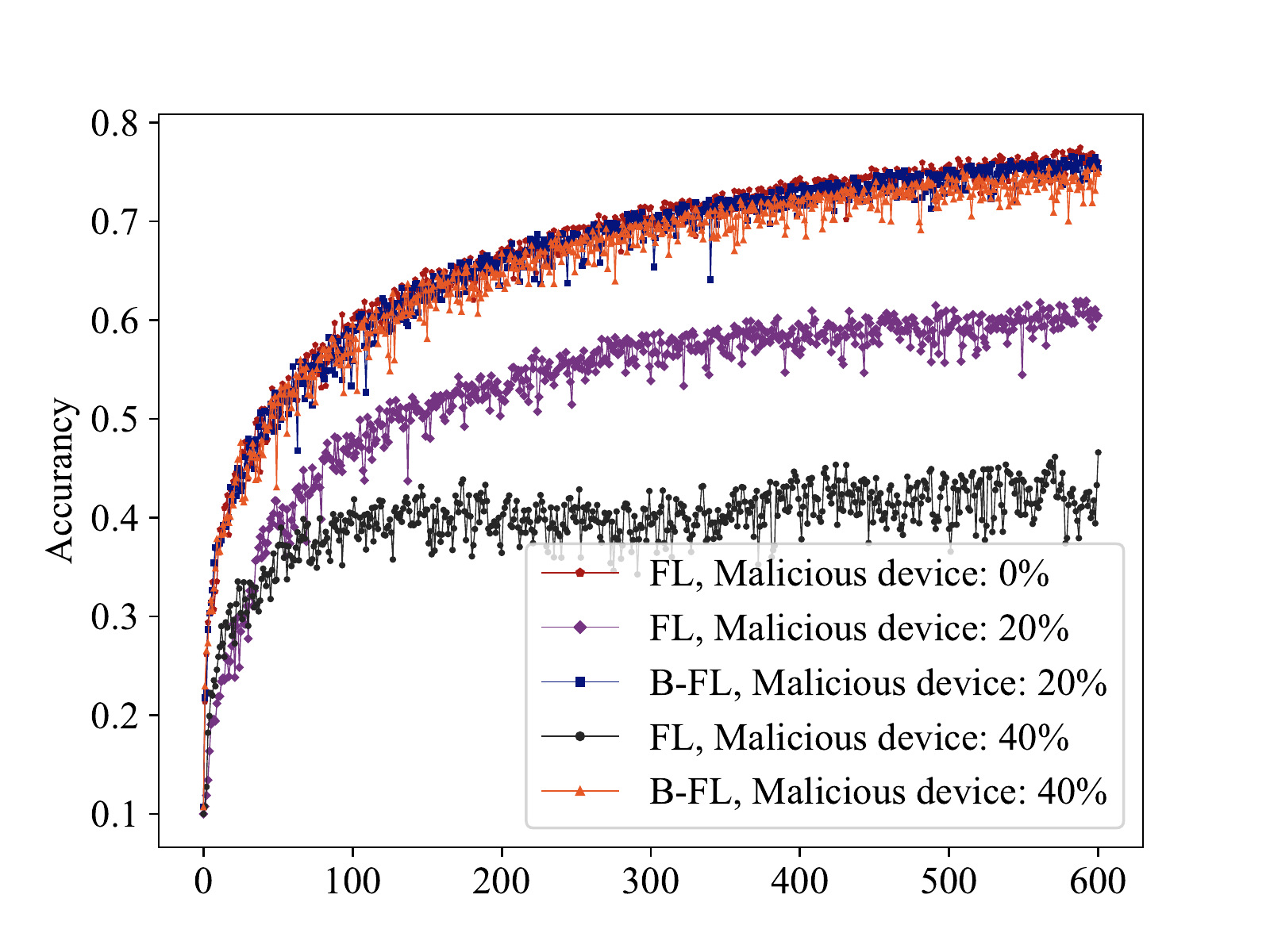}
		\caption{Test accuracy of FL and B-FL with malicious edge devices for CIFAR10 dataset.}\label{fig_accurancy_cifar10}
	\end{minipage}
\end{figure}

\subsection{Performance of wireless B-FL}
\label{sec_sim_B-FL}

We simulate the processes of B-FL following Algorithm \ref{alg_FL+blockchain}. The malicious edge devices is assumed to upload local models with random DNN parameter, which follow normal distribution $\mathcal{N}(0,1)$ and  will obviously  influence the convergence performance of FL training \cite{blanchard2017machine}. Besides, we use FedAvg algorithm as global model aggregation algorithm in FL training process \cite{yang2020federated}.

In the handwriting recognition task, we set that the percentage of  malicious edge devices  ranges from $0\%$ to $100\%$ and train CNN in 100 global epochs. The results are shown in Table \ref{tab_acc}, which shows the  malicious edge devices will obviously decrease the accuracy. The proposed B-FL in this paper shows robustness for the  malicious edge devices, which can not be affected below $50\%$. The partial  loss and accuracy of training rounds are shown in Fig. \ref{fig_accuracy_minist} and  Fig. \ref{fig_loss_minist}. The results show that the proposed B-FL with 40\% malicious devices and FL with 0\% malicious devices have the same performance, which is because the secure global aggregation based on multi-KRUM algorithm can eliminate partial local models from malicious devices. Therefore, the global model is not affected by malicious local models. Furthermore, the influence of the number of aggregated local models is inconspicuous in this simulation settings.

\reviseZP{In image classification task, we set the percentage of malicious edge devices  as $0\%,20\%$ and $40\%$. The test loss and test accuracy are shown in Fig. \ref{fig_loss_cifar10}, and Fig. \ref{fig_accurancy_cifar10}, which indicate that the malicious devices 
yield significant influence on the FL training process. However, the proposed B-FL with 40\%  malicious edge devices can still maintain the similar accuracy with FL with $0\%$ malicious edge devices, which is mainly because the proposed B-FL  utilizes secure global aggregation based on multi-KRUM algorithm to eliminate local models from malicious devices.}

In the affect recognition task, we assume the FL and B-FL systems have $10\%$ malicious edge devices to attack  global model respectively and present the  performance of centralized FL and proposed B-FL in different settings. The loss and accuracy of training rounds are shown in Fig. \ref{fig_loss_fig} and Fig. \ref{fig_accurancy_fig}. The proposed B-FL can conspicuously resist the attack of malicious edge devices and achieve excellent performance. We just plot the scenario of  $10\%$ malicious edge devices in this task, which is because the test loss is diverging when the number of  malicious edge devices continues to increase.

\begin{figure}[t]
	\centering
	\includegraphics[width=1\linewidth]{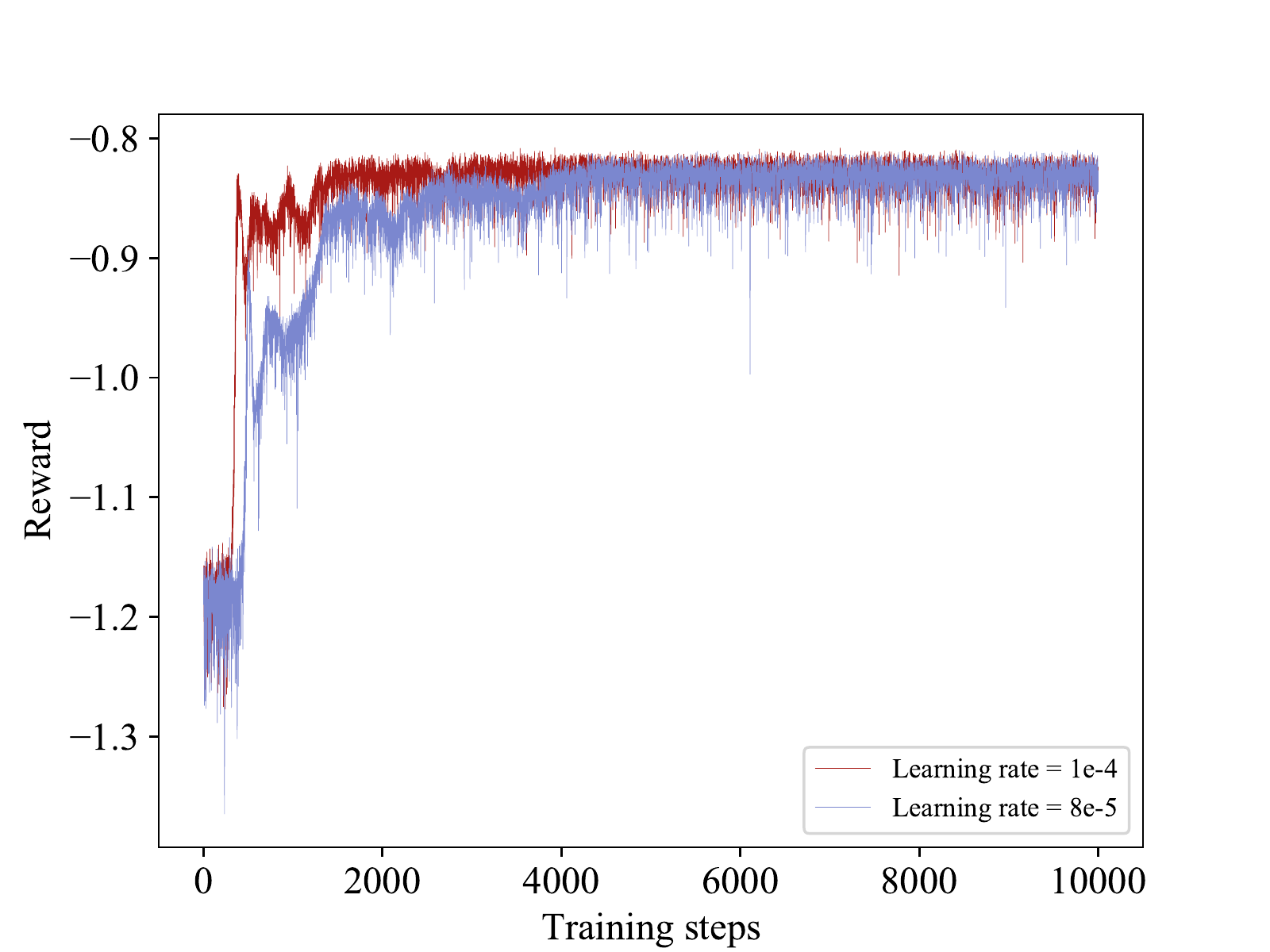} 
	\caption{Convergence performance of TD3-based resource allocation algorithm.}\label{fig_training}
\end{figure}

On the other hand, the situation of malicious edge servers is also considered in this paper. It's obvious that the FL training process will be discontinued or destroyed completely when the only edge server is malicious. However,  this will not happen in B-FL system, which has multiple edge servers to reach consensus and produce global model. When the primary edge server is malicious, the other validator edge servers will not accept its block and global model and will choose another new primary edge server. If less than $1/3$ validator edge servers are malicious, the  consensus can not be destroyed because of the mechanism of PBFT. If more than $1/3$ validator edge servers are malicious, the  consensus will be destroyed and the ML training process will be discontinued or destroyed completely. However, the situation of  more than $1/3$ malicious validator edge servers is almost impossible because the authorization is generally required for validator edge servers to participate in the consensus.

\subsection{Convergence of TD3 Based Resource Allocation Algorithm}
\begin{figure}[t]
	\centering
	\includegraphics[width=1\linewidth]{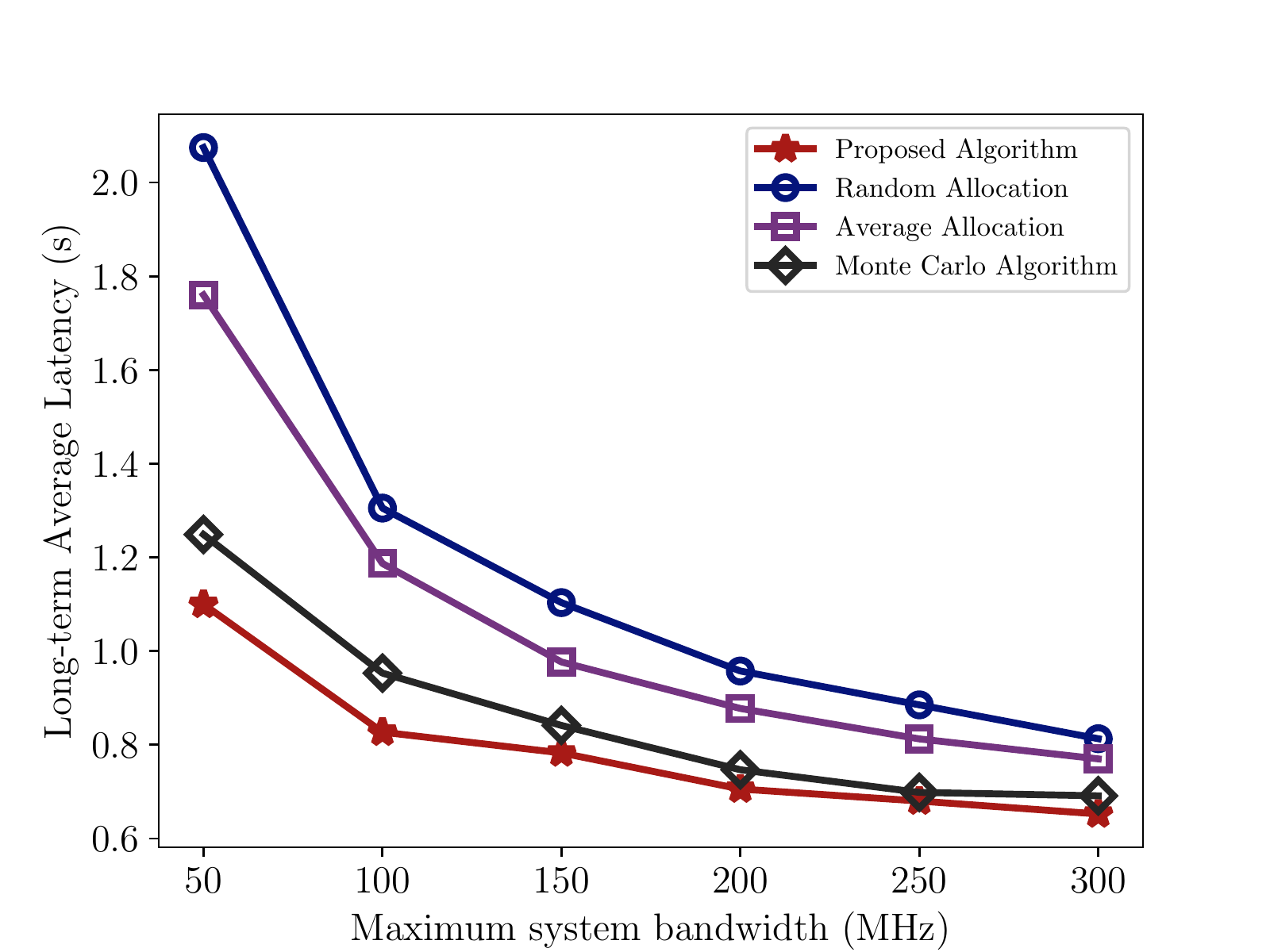} 
	\caption{Latency $(s)$ vs. maximum system bandwidth (MHz).}\label{fig_diff_B}
\end{figure}

\begin{figure}[t]
	\centering
	\includegraphics[width=1\linewidth]{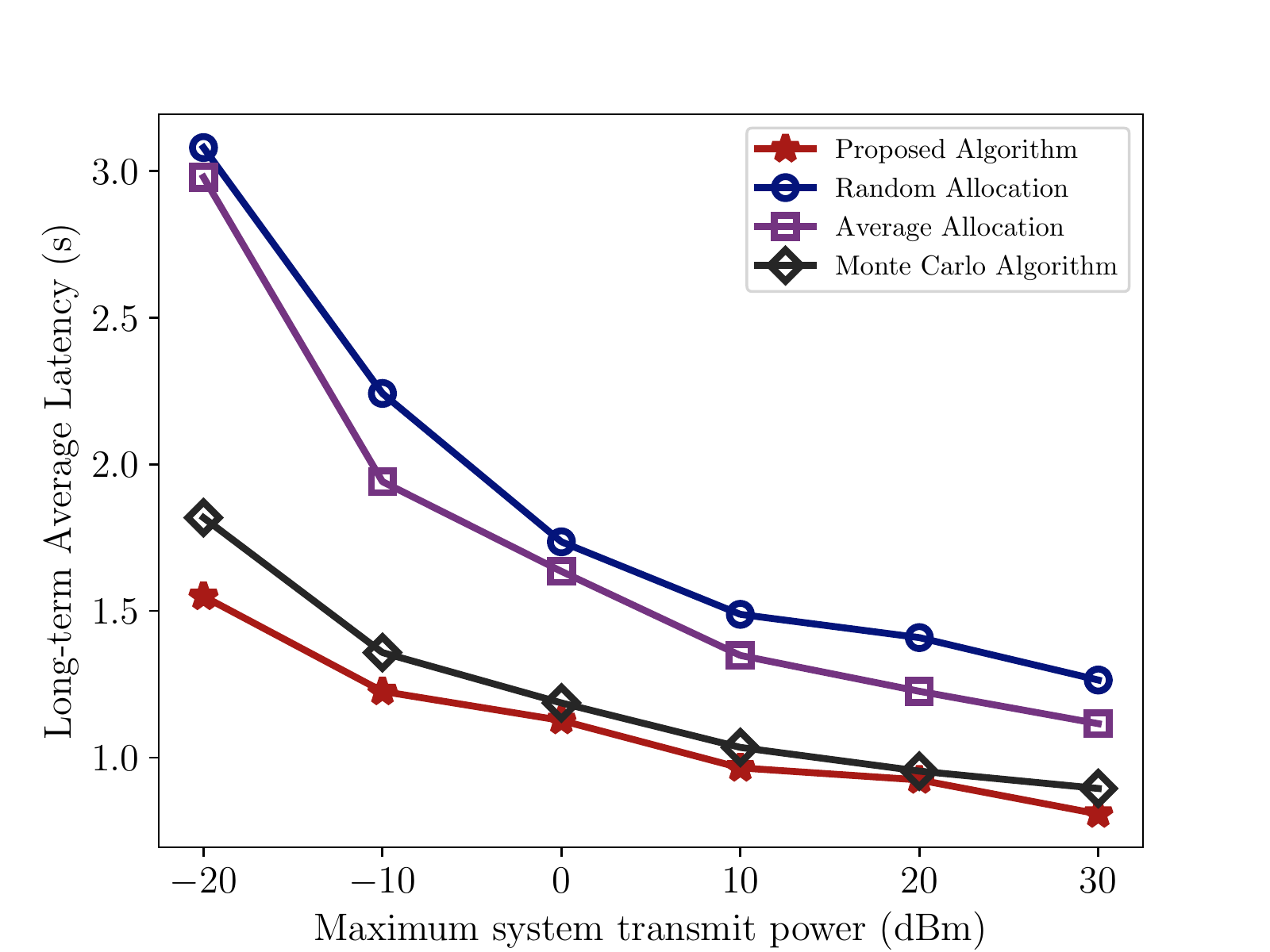} 
	\caption{Latency $(s)$ vs. maximum system transmit power (dBm).}\label{fig_diff_P}
\end{figure}

\begin{figure}[t]
	\centering
	\includegraphics[width=1\linewidth]{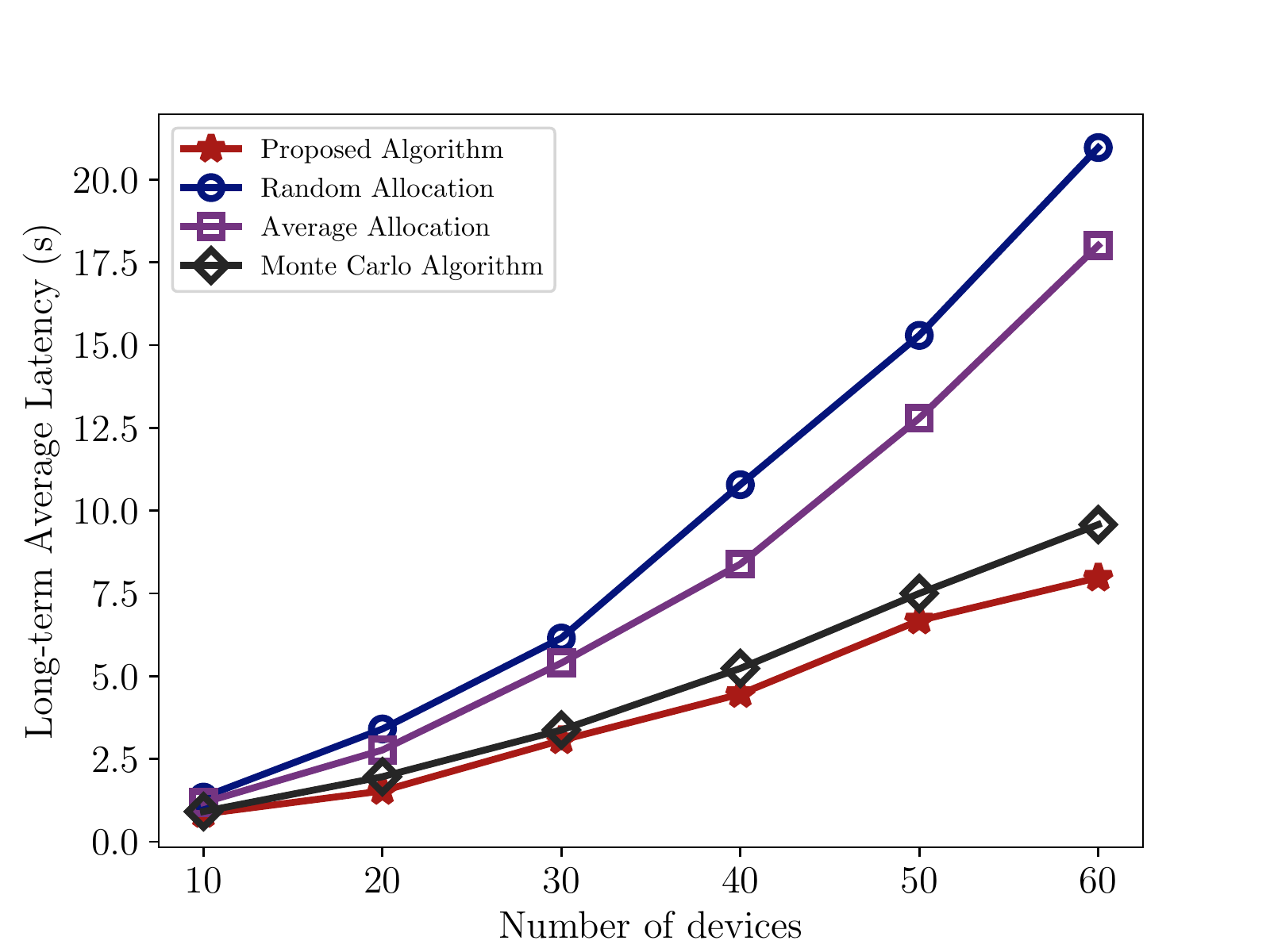} 
	\caption{Latency $(s)$ vs. number of edge devices.}\label{fig_diff_K}
\end{figure}

To evaluate  the effectiveness  of our proposed TD3 based
resource allocation algorithm, the convergence behavior of TD3 is shown in Fig. \ref{fig_training}. The performance is presented by instant rewards following the training steps.
We also illustrate the performance over different learning rate, i.e., $\eta_a=\eta_c=\{1\text{e}-4, 8\text{e}-5\}$. We choose the two learning rates because of the brief and clear  performance improvement. It can be seen that the rewards will converge with the increase of time step $t$. And the convergence rate will be promoted with  larger learning rate.

\subsection{Performance Impact of Different System Settings}

To evaluate the performance of proposed TD3 based resource allocation algorithm, we compare it with three benchmark algorithms, i.e., random allocation, average allocation, and Monte Carlo algorithm. Thus, we plot three figures to indicate the performance of our proposed algorithm and impact of different system settings, i.e., maximum system bandwidth, maximum system transmit power, and the number of edge devices.  \reviseZP{We average the results over 500 realizations of the system in our experiments to mitigate the randomness of devices' locations and channel states.}


Fig. \ref{fig_diff_B} illustrates the influence of maximum system bandwidth $b^{\max}$ to the long-term latency during B-FL training process.  Firstly, one observation is that our proposed TD3-based algorithm has the similar result with the benchmark Monte Carlo algorithm, and is better than the random allocation and average allocation schemes. 
\reviseZP{ In particular, the DRL-based algorithm is a little better than Monte Carlo algorithm, because the Monte Carlo method is computationally expensive and can only obtain a sub-optimal solution by using a finite number of samples.}
We take this result to present the performance of our TD3-based algorithm, which chooses more efficient and better resource allocation scheme. Secondly,  the long-term latency can significantly decrease with increasing maximum system bandwidth. Similarly, we present the impact of maximum system transmit power  in Fig. \ref{fig_diff_P}.

Fig. \ref{fig_diff_K}  plots the long-term latency to present the scalability to deal with different network scale of edge devices in the system. The number of edge devices varies from 10 to 60 and the other system settings are fixed. In the simulation, we find that TD3-based algorithm achieves greater gaps than three benchmarks with increasing number of edge devices. The long-term latency is  increasing over more edge devices, which is because the average resources decrease and the size of new block increase.

\reviseZP{   
	For practical implementations, the B-FL system can be implemented by using the open source blockchain frameworks, such as Ethereum, Hyperledger Fabric, Corda, and FISCO BCOS. Based on the blockchain framework, the local model can be submitted as a transaction and the global model aggregation and validation algorithms can be deployed as smart contracts. Particularly,  \cite{zhang2021blockchainbased} and \cite{kang2019incentive} have already deployed FL in the Ethereum framework and Corda framework, respectively.
          }

\section{Conclusion}

In this paper, we developed a B-FL architecture to ensure the security and privacy, which utilizes secure global aggregation and blockchain technique to resist the attacks from malicious edge devices and servers. We utilized PBFT consensus protocol in B-FL to achieve high effectiveness and low energy consumption for trustworthy FL. The procedures of PBFT-based wireless B-FL was presented at first, and the training latency was then analyzed. We have formulated an optimization problem that jointly considers bandwidth allocation and transmit power allocation to minimize the long-term average training latency. To solve this network optimization problem, we derived TD3-based algorithm to achieve long-term resource allocation and low computational complexity. Finally, we simulated the learning performance of wireless B-FL and efficiency of DRL based resource allocation algorithm, which is compared with baseline algorithms (i.e., random allocation, average allocation, and Monte Carlo algorithm). Our simulation results shown that the wireless B-FL architecture can resist the attacks from malicious servers and malicious devices. Furthermore, the training latency of wireless B-FL can be significantly reduced by the developed TD3-based adaptive resource allocation scheme.

\bibliographystyle{IEEEtran}
\bibliography{refs}

\end{document}